# scMEDAL for the interpretable analysis of single-cell transcriptomics data with batch effect visualization using a deep mixed effects autoencoder

Aixa X. Andrade[1], Son Nguyen[1], Austin Marckx[1] and Albert Montillo[1]*
[1]Lyda Hill Department of Bioinformatics
University of Texas Southwestern Medical Center
Dallas, TX 75390, USA
*Corresponding author: Albert Montillo[1], Ph.D.; Email: albert.montillo@utsouthwestern.edu

## Abstract

Single-cell RNA sequencing enables high-resolution analysis of cellular heterogeneity, yet disentangling biological signal from batch effects remains a major challenge. Existing batch-correction algorithms suppress or discard batch-related variation rather than modeling it. We propose **scMEDAL**—*single-cell Mixed Effects Deep Autoencoder Learning*—a framework that separately models batch-invariant and batch-specific effects using two complementary subnetworks.

The principal innovation, **scMEDAL-RE**, is a ***r****andom-****e****ffects Bayesian autoencoder* that learns batch-specific representations while preserving biologically meaningful information confounded with batch effects—signal often lost under standard correction. Complementing it, the ***f****ixed-****e****ffects subnetwork*, **scMEDAL-FE**, trained via adversarial learning provides a default batch-correction component.

Evaluations across diverse conditions (autism, leukemia, cardiovascular), cell types, and technical and biological effects show that scMEDAL-RE produces interpretable, batch-specific embeddings that complement both scMEDAL-FE and established correction methods (scVI, Scanorama, Harmony, SAUCIE), yielding more accurate prediction of disease status, donor group, and tissue.

scMEDAL also provides generative visualizations—including counterfactual reconstructions of a cell's expression as if acquired in another batch. The framework allows substitution of the fixed-effects component with other correction methods, while retaining scMEDAL-RE's enhanced predictive power and visualization. Overall, scMEDAL is a versatile, interpretable framework that complements existing correction, providing enhanced insight into cellular heterogeneity and data acquisition.

## 1. Introduction

Data from single-cell RNA sequencing has the potential to provide key new insights into cellular heterogeneity by capturing gene expression variations at single-cell resolution. However, challenges inherent to the data including data sparsity[1], high dimensionality[2], and batch effects[3], complicate its analysis. Batch effects, which can arise from biological variability (e.g., differences in donors or patients sampled, cell cycle stage, or tissue collection) or technical factors[4,5], (e.g., library preparation or sequencing platform) hinder the identification of biological relationships such as cell types and gene expression patterns.

As sequencing costs continue to fall, the volume and diversity of scRNA-seq data have expanded rapidly[6], intensifying the need for robust batch-effect correction methods to fully exploit these datasets. Existing approaches span several major method families. **Manifold and graph alignment** methods create linear embeddings using principal component analysis (PCA)[7], singular value decomposition (SVD), or canonical correlation analysis (CCA), then integrate datasets across batches through nearest-neighbor alignment (Harmony[8], Scanorama[9]) or anchor matching (Seurat[10]). **Deep learning autoencoders** instead learn flexible, non-linear latent spaces. For example, *scAlign* constructs a shared latent space via a round-trip random-walk objective; *Deep Count Autoencoder*[11] denoises counts without performing integration; *BERMUDA*[12] applies transfer learning with a maximum mean discrepancy loss to align similar clusters while preserving batch-specific signal; and *DESC*[13] iteratively self-trains embeddings to reduce batch effects without requiring batch labels. **Variational autoencoder (VAE)** models, such as *scVAE*[13], *scVI*[14], and *scANVI*[15] add incorporate probabilistic inference to capture uncertainty, with *scANVI* further enabling semi-supervised learning.

Building upon these foundations, **adversarial autoencoder methods**—both domain-adversarial and generative—promote **batch-invariant representations** by training an encoder (or generator) against a batch discriminator, as exemplified by *scGAN*[16]. These frameworks often combine multiple objective functions to refine batch correction. For instance, *iMAP* integrates adversarial learning with mutual nearest-neighbor (MNN) pairing to align datasets to an anchor batch[17], while *ResPAN* employs a residual autoencoder with skip connections, leveraging MNN alignment without relying on an anchor[18]. More recent methods extend adversarial strategies by incorporating **batch and cell-type classifiers** to improve supervision and integration. *scDREAMER* includes both batch and optional cell-type classifiers for supervised correction[19]; *IMAAE* uses an anchor batch GAN coupled with a cell-type classifier[20]; *Autoencoder-based Batch Correction (ABC)* applies dual cell-type classifiers to both latent and reconstructed spaces[21]; and *Dynamic Batching Adversarial Autoencoder (DB-AAE)* reduces training noise by dynamically varying batch sizes[22].

However, the aforementioned batch correction methods focus on removing batch effects without explicitly modeling them, potentially leading to overcorrection or loss of biological information—especially when signals are intertwined with batch effects[23]. Even though methods may incorporate cell-type labels (e.g., ABC, AutoClass[24] and scANVI) or impose similarity constraints (e.g., ResPAN, Adversarial Information Factorization (AIF[25])) to enhance cell-type signals, balancing effective batch correction with the retention of biological variation remains challenging. We hypothesize that characterizing batch variability will provide multiple benefits, including an improved understanding of scRNA-seq data and enabling accurate predictions of biological states at the cellular level, including diagnoses and cell-types. Therefore, this work develops a novel batch suppression framework, scMEDAL, which is the first to separately characterize and model batch-invariant and batch-specific variations using a mixed effects deep learning approach. This framework, called **scMEDAL** (***s****ingle-****c****ell* ***M****ixed* ***E****ffects* ***D****eep* ***A****utoencoder* ***L****earning*), is inspired by the mathematical foundation of statistical mixed effects models —with their fixed and random components—and is likewise implemented using two complementary deep autoencoder subnetworks that capture fixed and random effects. One network is based on adversarial deep learning and learns the batch-invariant scRNA-seq representation (i.e., the fixed effects), while the other is a *Bayesian* neural network autoencoder which learns the batch-specific scRNA-seq probabilistic representation (i.e., the random effects) through variational inference. Unlike other batch correction methods (e.g., iMAP[17], ResPAN[18], scDREAMER[19], IMAAE[20], ABC[21], DB-AAE[22]) that focus on creating batch-invariant spaces without including a random-effects subnetwork or explicitly modeling batch distributions, scMEDAL explicitly learns batch distributions, which we show are necessary for recovering meaningful signals that might otherwise be discarded by batch-invariant only approaches.

The scMEDAL incorporates multiple visualization modules designed to enhance interpretability of its learned representations. One component generates low-dimensional visualizations of the latent embedding spaces using Uniform Manifold Approximation and Projection (UMAP), enabling qualitative assessment of how well biological variation and batch effects are captured or separated. To evaluate performance across a broad range of scenarios, we evaluated across diverse single-cell and single-nucleus RNA-seq datasets spanning multiple health conditions, cell types, and batch effects. These datasets include data from cardiovascular health (Healthy Heart), Autism Spectrum Disorder (ASD), and Acute Myeloid Leukemia (AML), spanning over 50 different cell types, and incorporating technical batch effects (e.g., sequencing platform and library preparation) and biological batch effects (e.g., disease status and donor variability).

The primary innovation of scMEDAL is its **random-effects module** (**scMEDAL-RE**), which explicitly models and visualizes batch-specific variability. By learning these batch-dependent representations, scMEDAL-RE produces **batch-enhanced embeddings** that reveal how technical or biological batch factors influence gene expression. This approach allows scMEDAL to **preserve biologically meaningful variation**—such as differences associated with disease status, donor group, or tissue type—that is often removed or obscured by conventional batch-correction methods, thereby enhances the interpretability of downstream analyses.

By leveraging the generative nature of the framework's fixed- and random-effects autoencoder networks, scMEDAL can answer retrospective, "What if?" analyses including predicting a cell's expression under different batches through genomap projections at the cellular level. This reveals the impact of biological batch effects (e.g., the diagnosis) on cellular heterogeneity and the impact of technical batch effects on the gene expression profile, as well as the potential impact of further steps taken to control such variability.

Because the subnetworks are trained independently, the **batch-specific latent space** learned by scMEDAL-RE can be seamlessly combined with **batch-agnostic embeddings** from existing unsupervised, self-supervised, or supervised correction methods. Integrating these complementary representations consistently improves the prediction of biological features that are confounded with batch—such as disease status, donor group, or tissue type—yielding greater classification accuracy than any single batch-invariant embedding alone.

# 2. Results

## 2.1 Overview of the scMEDAL framework: Capturing fixed and random effects in scRNA-seq data

The architecture of the scMEDAL framework, illustrated in **Fig**. 1, is designed for single-cell RNA sequencing (scRNA-seq) data. scMEDAL addresses batch effects—technical or biological variations that can confound analyses—by separately modeling and quantifying the fixed and random effects across the gene expression vectors of the cells. Fixed effects represent batch-invariant features, while random effects capture batch- or donor-specific variations. By separately modeling both types of effects, scMEDAL decouples batch effects from the data, enhancing explainability and preserving biologically meaningful information entangled with batch. The scMEDAL framework processes a gene expression count matrix $\boldsymbol{X} \in \mathbb{R}^{n \times g}$, where $n$ is the number of cells and $g$ is the number of genes. The framework learns two lower dimensional latent space representations: one encodes batch-agnostic fixed effects, referred to as the FE latent space representation, and the other captures batch-specific random effects, called the RE latent space representation. The architecture of our framework includes two parallel subnetworks, each of which is a type of autoencoder. (1) The *Fixed Effects subnetwork (**scMEDAL-FE**)* suppresses batch effects by learning a batch-invariant latent representation of the gene expression data. Its encoder and decoder each have two dense hidden layers with weight tying and are guided by the addition of an adversarial classifier that aims to predict the batch label, $\boldsymbol{z}$ (**Fig.** 1) using the intermediate outputs from the encoder layers. To the extent that it can, this classifier penalizes the encoder, thereby encouraging the learning of an embedding that is not predictive of the batch but is still informative for gene expression reconstruction. (2) The *Random Effects subnetwork (**scMEDAL-RE**)*, our principal innovation, is a *Bayesian* autoencoder, which learns a batch-specific representation. A Bayesian network replaces the traditional point estimate weights with weight distributions to model the RE as random variables. It is efficiently trained via variational inference. This generative model can be used as a visualization tool to understand how gene expression profiles vary across batches and thereby provide further insight into their technical or biological underpinnings.

scMEDAL-RE enables more accurate modeling of biological signals while mitigating the information loss typically incurred by batch correction methods that entirely suppress batch effects. To assess scMEDAL-RE's complementary nature, we use a Random Forest classifier[26] as a downstream mixing classifier to predict tissue, diagnosis, patient group, and cell-type labels. We extend this classification pipeline to demonstrate that scMEDAL-RE recovers biological information previously entangled with batch effects not only for our default scMEDAL-FE subnetwork but also for alternative dimensionality-reduction methods such as Principal Component Analysis (PCA)[7], and, batch correction approaches such as scVI[14], Harmony[8], Scanorama[9], SAUCIE[27]. Additionally, a variation of our default batch-invariant subnetwork is developed to exploit cell-type labels when they are available. This variation replaces the autoencoder network with an autoencoder classifier, which we denote as the *Fixed Effects Classifier subnetwork* (***scMEDAL-FEC***), and which predicts cell type to further enhance the cell-type signal. Finally, an additional comparable, we include the cell type-label supervised scANVI[15] and demonstrate that scMEDAL-RE's benefits extend beyond our default batch-invariant subnetworks to alternative supervised approaches.

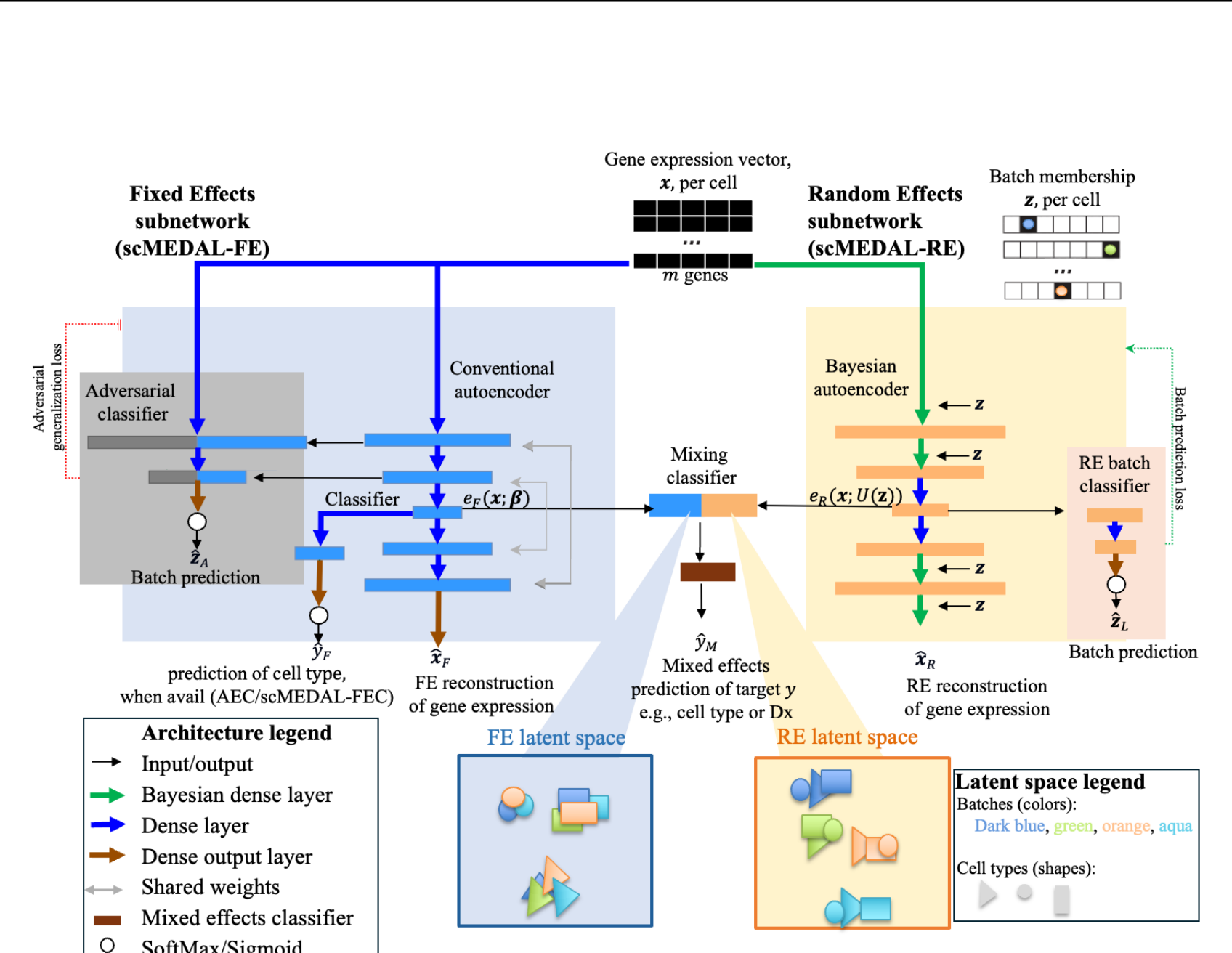

**Fig. 1. Overview and architecture of the scMEDAL framework**. The inputs to the model are the gene expression vector $\boldsymbol{x}$ and the one-hot encoded batch vector $\boldsymbol{z}$ per cell. Both the fixed- and random-effects subnetworks aim to learn a latent representation of the input data which can be used to produce a reconstruction $\hat{\boldsymbol{x}}$ of the input gene expression. The *Fixed Effects subnetwork (scMEDAL-FE)* shown in the blue block, comprises a conventional autoencoder augmented with an adversarial classifier (gray block), which attempts to predict $\boldsymbol{z}$ using features from the AE. To the extent that it can, this classifier penalizes the conventional AE for using such features, encouraging the AE to learn a batch-invariant latent space that preserves reconstruction ability and, hence, cell-type signal $y$. When cell-type labels are available, we use an autoencoder classifier (AEC)—adding an auxiliary classifier to the AE to form scMEDAL-FEC—to further strengthen cell-type structure in the embeddings. The *Random Effects subnetwork (scMEDAL-RE)* shown in the orange block, includes an autoencoder (orange AE, left) and a parallel MLP classifier designed to predict the batch, $\boldsymbol{z}$. This MLP encourages the scMEDAL-RE to capture batch variability in its latent representation by modeling the effects associated with each batch. Overall, the fixed-effects latent space representation is batch-invariant, while the random-effects latent space captures and enhances batch-specific variability. Both scMEDAL-FE and scMEDAL-RE are trained separately to achieve better convergence. After training both scMEDAL subnetwork, a mixing classifier depicted in brown, integrates these latent spaces to predict the labels of interest $\hat{y}$ (e.g., diagnosis or cell type). This enables an assessment of how well the complementary latent spaces preserve biological signals for accurate cell type classification.

To evaluate cell type separability, batch correction, and batch modeling capabilities, we adopt three clustering metrics: the Average Silhouette Width (ASW)[28-30], Calinski-Harabasz (CH) index[29-32], and Davies-Bouldin (DB) index[29,33]. ASW is our primary metric because its values are bounded from -1 to 1, and it is widely used in scRNA-seq analysis[19,23,34-36], facilitating comparisons across methods and datasets. Additionally, ASW is the most robust of the three metrics, as it effectively captures variations in cluster density, shape, and size. DB is included as it excels at identifying well-separated clusters, while CH is particularly useful for evaluating clustering structure when the number of clusters is well-defined[37].

Once the latent spaces are learned, UMAP[38] is used to assess qualitatively and in an interpretable manner, the extent to which the spaces preserve biologically meaningful information. These projections complement the quantitative metrics discussed above. Additionally, genomaps are used to provide a 2D visualization of individual cell gene expression based on gene-gene interactions. The scMEDAL framework is comprehensively evaluated using three diverse single-cell and single-nucleus RNA-seq datasets, spanning multiple disease conditions, a wide variety of cell types and batch variations. This assessment includes datasets from the cardiovascular system (Healthy Heart), ASD, and AML (Supplementary **Table** S1). Data preprocessing follows standard scRNA-seq practices[36] including UMI count filtering and normalization, log transformation, and selection of highly variable genes. The datasets are then partitioned using five-fold cross-validation, stratified by batch and cell type (see Data preprocessing section 2.2 in the Supplementary information).

## 2.2 scMEDAL creates complementary batch-invariant and batch-specific latent spaces in the Healthy Heart dataset

The first dataset, the *Healthy Heart dataset*[36,39] consists of scRNA-seq data from tissues of the heart from healthy individuals. It contains multiple sources of batch variability, including two different sequencing protocols (technical effects), samples from various donors, and different tissues within the heart (biological effects). This complexity makes it a rigorous benchmark for evaluating scMEDAL. We compared scMEDAL against widely used unsupervised methods for preprocessing scRNA-seq data, including Principal Component Analysis (PCA)[7] and the Autoencoder (AE) neural network[40], scVI, Harmony, Scanorama, and SAUCIE.

To **qualitatively** compare the latent spaces learned by these approaches, we apply UMAP[38] to visualize the distribution of 41544 cells from each model's latent space(s) (**Fig.** 2a and b). scMEDAL-RE stands out as the only method that cleanly separates all sources of batch effects—protocol, tissue, and donor—while other methods do not (Fig. 2a, fourth column; Supplementary Fig. S3). PCA and AE retain some batch structure, but less distinctly, and all other batch-correction methods increase visible mixing across protocols and donors compared with scMEDAL-RE. Crucially, tissue separation is preserved only in scMEDAL-RE (Fig. 2a, third row), highlighting its ability to recover biologically meaningful signals that are typically lost during batch correction.

Cell-type separability, shown in the first row of Fig. 2a and 2b, is apparent in most batch correction models, including scMEDAL-FE, scVI, and Harmony, which produced more cohesive and well-defined clusters than PCA and AE. When colored by batch (second row of Fig. 2a and 2b), scVI, Harmony, and scMEDAL-FE displayed the most uniform distributions, while PCA, AE, and Scanorama retained noticible batch structure. Some clusters remained unresolved across all methods—for example, a persistent "Not Assigned" cell type[41] cluster in scMEDAL-FE arising from atrial cells spanning two protocols but only two batches out of the 20 shown (147 in total), giving them lower representation than other tissue–protocol combinations. Moreover, atrial and ventricular cardiomyocytes merged in most correction methods (except Scanorama), likely reflecting over-suppression of tissue-level variability. By contrasting the AE and scMEDAL-FE representations, we identified batch-invariant cell types, further clarifying cell variability.

A **quantitative** analysis confirmed these qualitative findings (Fig. 2c). First, we evaluated batch separability. Here scMEDAL-RE attained an ASW of +0.10, substantially modeling batch distributions, while scMEDAL-FE suppressed batch information (ASW −0.21). PCA and AE provided baselines ASW of −0.18 and -0.17, respectively (**Fig.** 2c, left columns). Additional clustering metrics (CH and DB,

Supplementary **Table** S6) further corroborate the efficacy of the proposed approach. These results demonstrate that unlike AE or standard correction methods, scMEDAL reduces dimensionality while both suppressing and modeling batch effects.

Next, we evaluated **cell-type separability**. Here we observe a wide range of performance across the methods, from the lowest performing method, SAUCIE (ASW -0.06) to the best performing method, scVI (0.22), with several methods attaining a middle performance (ASW 0.10-0.15). (**Fig.** 2c, right columns; Supplementary **Table** S7).

Collectively, these qualitative and quantitative results highlight how scMEDAL-RE uniquely enables high-resolution visualization of batch information while maintaining biologically relevant tissue signals, representing a complementary advance to existing batch correction methods. Additionally, scMEDAL-FE, our default batch-invariant component, can be replaced by any other batch-invariant method that does not explicitly model batch effects (e.g., scVI), while retaining *tissue* information, which is demonstrated solely by scMEDAL-RE.

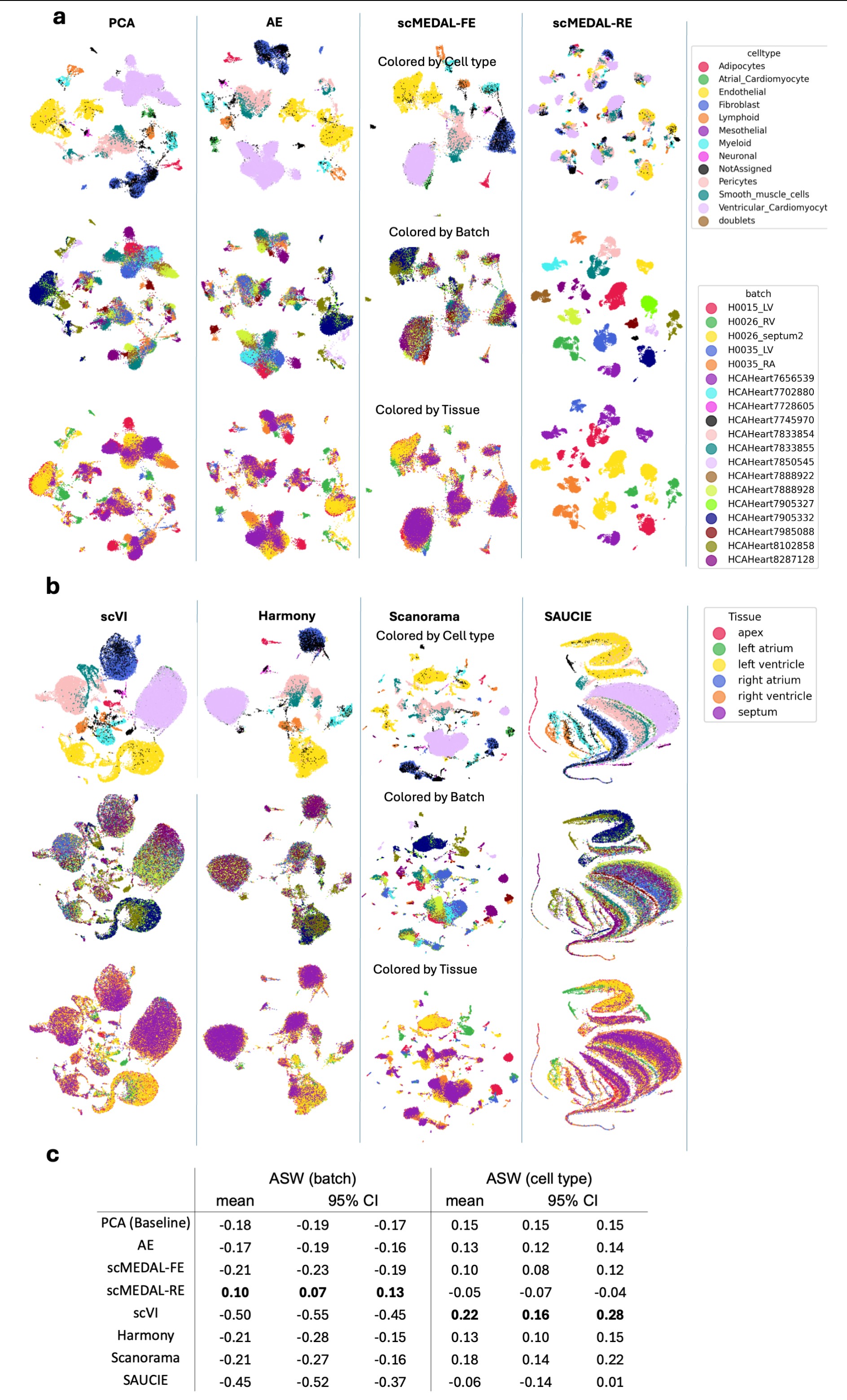

| | ASW (batch) | | | ASW (cell type) | | |
|---|---|---|---|---|---|---|
| | mean | 95% CI | | mean | 95% CI | |
| PCA (Baseline) | -0.18 | -0.19 | -0.17 | 0.15 | 0.15 | 0.15 |
| AE | -0.17 | -0.19 | -0.16 | 0.13 | 0.12 | 0.14 |
| scMEDAL-FE | -0.21 | -0.23 | -0.19 | 0.10 | 0.08 | 0.12 |
| scMEDAL-RE | **0.10** | **0.07** | **0.13** | -0.05 | -0.07 | -0.04 |
| scVI | -0.50 | -0.55 | -0.45 | **0.22** | **0.16** | **0.28** |
| Harmony | -0.21 | -0.28 | -0.15 | 0.13 | 0.10 | 0.15 |
| Scanorama | -0.21 | -0.27 | -0.16 | 0.18 | 0.14 | 0.22 |
| SAUCIE | -0.45 | -0.52 | -0.37 | -0.06 | -0.14 | 0.01 |

**Fig. 2: scMEDAL subnetworks output complementary batch-invariant and batch-specific latent spaces in the Healthy Heart dataset. a** UMAP visualization of 41544 cells from 20 out of 147 batches of the Healthy Heart dataset's latent spaces, generated using the following models: PCA (first column), AE (second column), scMEDAL-FE (third column) and scMEDAL-RE (fourth column). **b** scVI (first column), Harmony (second column), Scanorama (third column) and SAUCIE (fourth column). **c** Average silhouette width (ASW) scores (mean and 95% confidence interval across 5 folds) for batch and cell-type separability in the latent spaces, using the PCA, AE, scMEDAL-FE, scMEDAL-RE, scVI, Harmony, Scanorama and SAUCIE models. Bolded scores indicate the highest separability for each metric.

## 2.3 scMEDAL-RE preserves disease-associated neuronal patterns in ASD

The second dataset, the *Autism Spectrum Disorder (ASD)*[42], includes single-nucleus (sn) RNA-seq data from brain samples of the prefrontal cortex (PFC) and anterior cingulate cortex (ACC). Batch effects primarily arise from donor variability between ASD and typically developing (control) subjects. There is indirect confounding between cell type and batch, as certain cell types are more affected in ASD than in controls. This dataset enables us to evaluate scMEDAL's capacity to capture donor heterogeneity and disease-related biological variations.

Qualitatively, we see in Fig. 3a (fourth column), that scMEDAL-RE (second and third rows) distinctly separates donors (colored batches) while preserving the diagnostic signal (red/green colors). ASD–control gene expression differences remain evident, unlike in all other methods where donor and diagnostic signals are completely intermixed. We also observe that, across all methods, certain cell types appear more dispersed than others, likely reflecting the donor-to-donor variability in cell type heterogeneity. The greatest dispersion is observed in L2/3, L4, L5/6, L5/6-CC, and Neu-mat populations; except for Neu-mat, these correspond to the most commonly affected cell types in ASD, as reported by Velmeshev et al., 2019[43]. We also observe that all batch correction methods, including scMEDAL-FE, show strong merging of astrocyte populations AST-FB (green color) and AST-PP (red color), which is expected given the donor confounding in these cells.

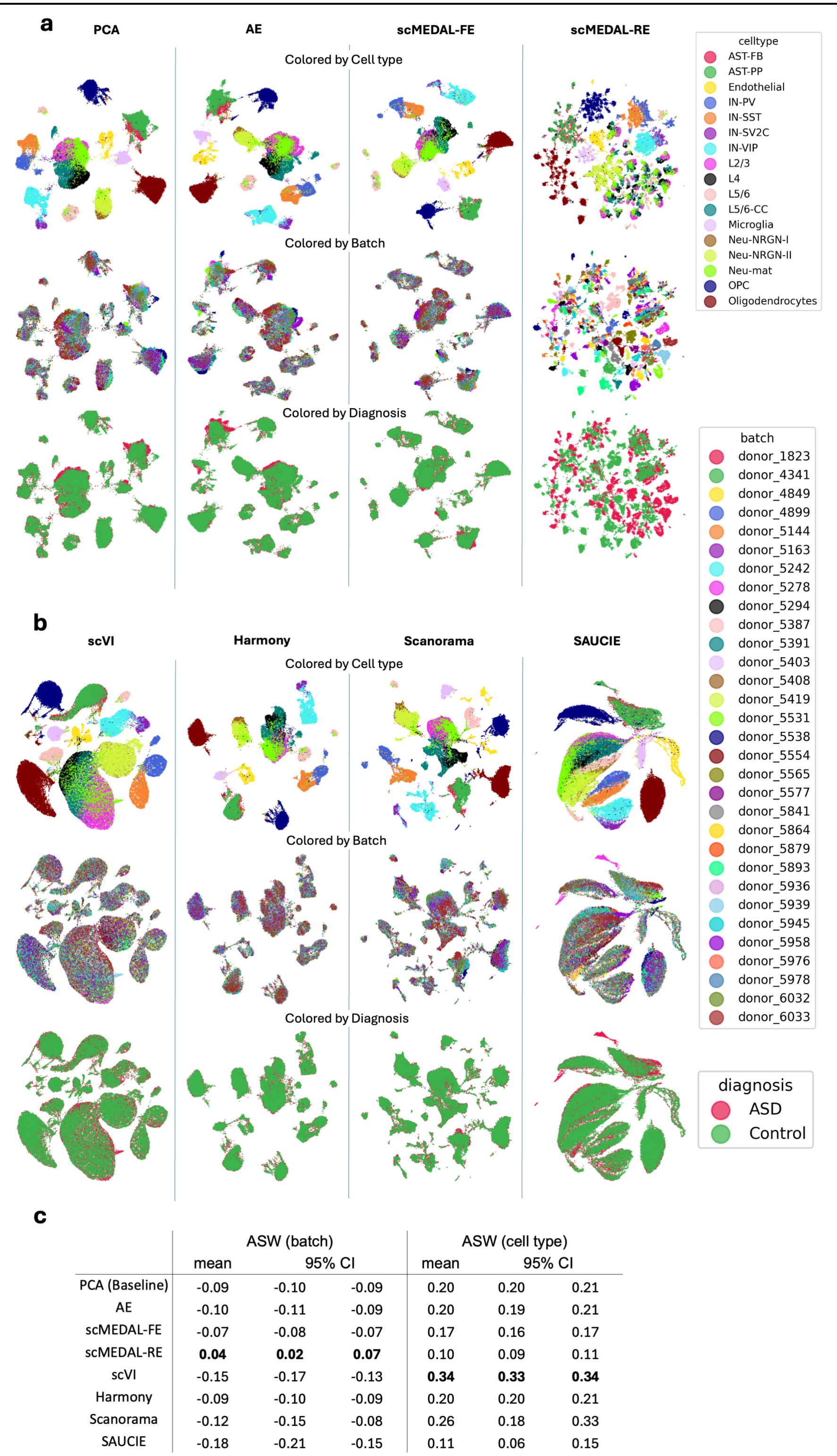

| | ASW (batch) | | | ASW (cell type) | | |
|---|---|---|---|---|---|---|
| | mean | 95% CI | | mean | 95% CI | |
| PCA (Baseline) | -0.09 | -0.10 | -0.09 | 0.20 | 0.20 | 0.21 |
| AE | -0.10 | -0.11 | -0.09 | 0.20 | 0.19 | 0.21 |
| scMEDAL-FE | -0.07 | -0.08 | -0.07 | 0.17 | 0.16 | 0.17 |
| scMEDAL-RE | **0.04** | **0.02** | **0.07** | 0.10 | 0.09 | 0.11 |
| scVI | -0.15 | -0.17 | -0.13 | **0.34** | **0.33** | **0.34** |
| Harmony | -0.09 | -0.10 | -0.09 | 0.20 | 0.20 | 0.21 |
| Scanorama | -0.12 | -0.15 | -0.08 | 0.26 | 0.18 | 0.33 |
| SAUCIE | -0.18 | -0.21 | -0.15 | 0.11 | 0.06 | 0.15 |

**Fig. 3: scMEDAL subnetworks disentangle ASD-related donor effects. a** UMAP visualizations of latent spaces derived from 62,735 cells from 31 donors in the ASD dataset, obtained using the following models: PCA (first column), AE (second column), and scMEDAL-FE (third column) and scMEDAL-RE (fourth column). **b** scVI (first column), Harmony (second column), Scanorama (third column) and SAUCIE (fourth column). **c** Average Silhouette Width (ASW) scores (mean and 95% CI across 5 folds) for batch and cell-type separability in the latent spaces, comparing PCA, AE, scMEDAL-FE, scMEDAL-RE, scVI, Harmony, Scanorama and SAUCIE models. Bolded scores indicate the highest separability for each metric.

Quantitatively, scMEDAL-RE achieves the highest ASW for batch separability (+0.04; Fig. 3b, bottom row), confirming its ability to capture donor effects as intended. Importantly none of the other methods have a positive batch ASW. Additional clustering metrics (CH and DB; Supplementary Table S7) further support these findings. By contrast, scVI yields the lowest batch ASW (−0.15) but the highest cell-type ASW (0.34), making it the strongest alternative to our default scMEDAL-FE. However, scVI does not preserve diagnosis information, highlighting scMEDAL-RE's complementary value for visualizing donor and disease effects.

## 2.4 scMEDAL balances the trade-off between batch correction and preserving cell type information in leukemia

The third dataset, the *Acute Myeloid Leukemia (AML)*[44], presents one of the most challenging scenarios for disentangling cell type signal from batch heterogeneity. It comprises scRNA-seq data from AML-confirmed patients, healthy control individuals, and leukemia cancer cell lines. Batch effects stem from donor variability, including differences between healthy and AML subjects, as well as cell lines. This dataset features strong confounding between cell type and batch: malignant cells are only present in diseased subjects and cell lines, while healthy cells originate from both diseased and healthy patients (Supplementary Fig. S2). Such confounding makes this dataset an especially stringent test of any method aiming to separate batch from biology. In this setting, scMEDAL-RE provides a clear advantage. In Fig. 4a (fourth column), it uniquely separates (in the second row), the donors which is the batch variable, while preserving critical biological structure—distinguishing patient from cell-line samples and, crucially, AML from control cells (third row). Competing approaches fail to recover these separations, with a high degree of visible mixing of AML and controls and often the cell line as well. Furthermore, scMEDAL-RE show similar donors cluster tightly together, capturing heterogeneity due to differences in cell-type composition (e.g., presence or absence of T cells in AML donors).

Quantitatively, scMEDAL-RE successfully captures batch variability, with a substantially increased ASW of +0.69 (**Fig.** 4c, left columns). Results using additional clustering metrics, CH and DB, further confirm the superior batch separability provided by scMEDAL-RE's latent space (see Supplementary **Table** S8). All batch correction methods suppress batch effects at the cost of conflating biological structure which reflects in a lower ASW for cell type than the baselines PCA and AE which do not perform batch correction (**Fig.** 4c, right columns). See MUTZ3 separability (brown) in Fig 4.

Importantly, the AML dataset highlights why scMEDAL-RE is needed: batch correction approaches remove not only donor effects but also meaningful biological variation. Because malignant cell types are inherently confounded with donor identity, eliminating batch information inevitably erases some disease-related signal (Fig. 4c, right columns). scMEDAL-RE overcomes this limitation by explicitly modeling batch as a random effect, preserving biologically relevant structure while still disentangling donor-driven variability.

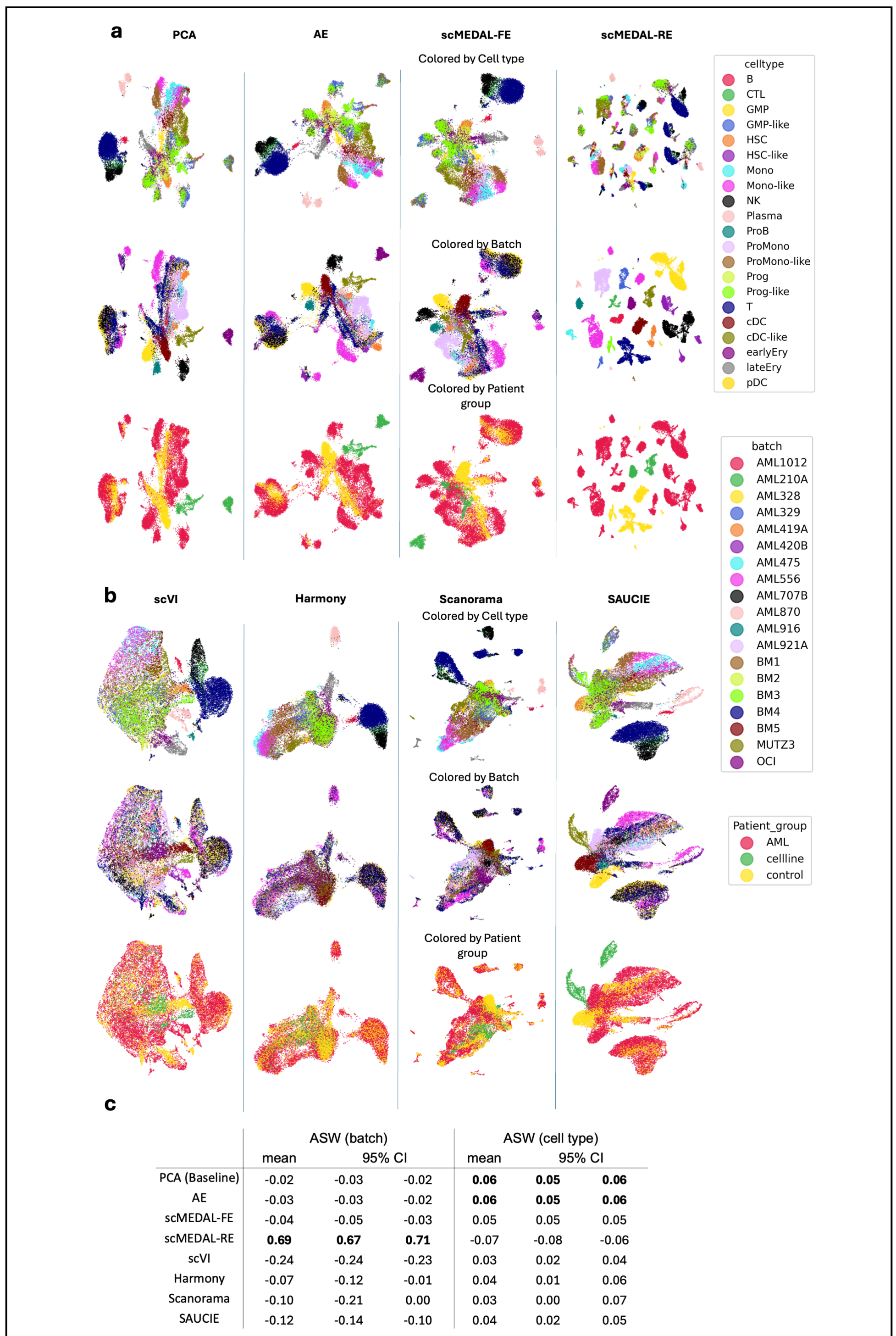

| | ASW (batch) | | | ASW (cell type) | | |
|---|---|---|---|---|---|---|
| | mean | 95% CI | | mean | 95% CI | |
| PCA (Baseline) | -0.02 | -0.03 | -0.02 | **0.06** | **0.05** | **0.06** |
| AE | -0.03 | -0.03 | -0.02 | **0.06** | **0.05** | **0.06** |
| scMEDAL-FE | -0.04 | -0.05 | -0.03 | 0.05 | 0.05 | 0.05 |
| scMEDAL-RE | **0.69** | **0.67** | **0.71** | -0.07 | -0.08 | -0.06 |
| scVI | -0.24 | -0.24 | -0.23 | 0.03 | 0.02 | 0.04 |
| Harmony | -0.07 | -0.12 | -0.01 | 0.04 | 0.01 | 0.06 |
| Scanorama | -0.10 | -0.21 | 0.00 | 0.03 | 0.00 | 0.07 |
| SAUCIE | -0.12 | -0.14 | -0.10 | 0.04 | 0.02 | 0.05 |

**Fig. 4: scMEDAL-RE captures patient group information complementing batch correction methods. a** UMAP visualization of latent spaces derived from 23,049 cells and 19 batches in the AML dataset. Control bone-marrow batches carry the BM prefix; the two cell lines are MUTZ3 and OCI. Columns show embeddings from PCA (first column), AE (second column), and scMEDAL-FE (third column) and scMEDAL-RE (fourth column). **b** scVI (first column), Harmony (second column), Scanorama (third column) and SAUCIE (fourth column). **c** Average Silhouette Width (ASW) scores (mean and 95% confidence intervals across 5 folds) for batch and cell-type separability in the latent spaces, comparing PCA, AE, scMEDAL-FE, scMEDAL-RE, scVI, Harmony, Scanorama and SAUCIE models. Bolded scores indicate the highest separability for each metric.

## 2.5 scMEDAL enables deeper understanding of batch effects by addressing retrospective “What if?” questions through generative modeling and visualization

The proposed scMEDAL batch suppression framework offers novel capabilities to address retrospective "What if?" questions. For example, how would a cell's gene expression profile change if it had originated from a different batch? Or, “How would the profile change if it came from a donor with a different diagnosis?”, or “from another donor with the same diagnosis?” Other questions include, “How would the profile change, if the cell came from a cell line rather than a patient?” This section demonstrates the framework’s ability to answer these questions.

To highlight the generative capabilities of the scMEDAL framework in addressing critical biological and clinical "What if?" questions, unexplored by traditional methods, we visualize gene expression patterns learned by the scMEDAL-RE subnetwork and contrast those with the batch independent patterns learned by a network such as the scMEDAL-FE. Both subnetworks are used to reconstruct gene expression count matrices. The FE subnetwork transforms input data into a batch-invariant space, revealing gene expression patterns common across all batches. Conversely, the RE subnetwork captures batch-specific variations, enabling the simulation of gene expression profiles as they would appear if cells originated from a different technical or biological batch.

By intentionally setting the one-hot encoded batch vector input to the RE subnetwork to a chosen batch, we can project cells onto that batch showing how its conditions distinct from their original batch (e.g., technical variations, different donors, or diagnoses). This allows exploration of batch variability across donors, diagnoses, tissues, or technical acquisition conditions.

Previous work (Islam *et al.*, 2023)[45] proposed **genomaps** as an image-based depiction of a cell’s gene expression pattern and demonstrated that they reveal clear differences between cell types. Specifically, a genomap is a 2D image, with each pixel representing a gene and its intensity corresponding to the gene’s expression level. The genes (pixels) are arranged in a polar plot, where genes with equivalent gene-to-gene interactions are positioned at the same distance from the image’s center (i.e., its origin with radius $\rho = 0$), and the angle $\theta$ is arbitrary. The mapping between the gene to its location in the polar plot is computed by the genomap transform[45].

Since no absolute benchmarking standard yet exists for genomap-based visualization, we focus on identifying biologically meaningful similarities and contrasts in the gene expression patterns captured by scMEDAL. That is, we compare genomaps for a single cell across batches of interest to enable visual inspection of the effects of how technical or biological batch factors influence expression profiles, while maintaining the same underlying biological signal.

To facilitate ready image to image comparison, we build a consistent genomap transformation that we use across all projections, ensuring consistent gene-to-pixel mappings for direct comparison, at the gene (pixel) level. To build this transform, we randomly select 300 cells from specific cell types of interest, while including cells from each batch of interest for the previously described “What-if?” questions. Using the scMEDAL-FE and RE subnetworks, we project their gene expression data. Counterfactual projections of each cell into all batches are generated with scMEDAL-RE. These projections (row vectors) are concatenated (stacked vertically) with the original gene expression vectors to form an overall count matrix, which is standardized. From this matrix, we compute a genomap transform[45] and apply it consistently to visualize gene expression as an image.

***Healthy Heart Dataset***

In the Healthy Heart dataset, to evaluate our method's effectiveness in learning a generative model of batch effects, we focus on two cell types: endothelial and myeloid cells—across four batches. These cell types and batches were chosen based on Fig. 2 and Supplementary Fig. S3, as each includes at least two batches generated using the same protocol. For each cell type, we select a cell at random to represent the cell type and visualize its gene expression using a genomap (**Fig.** 5, left column). By projecting these cells into different batches, as illustrated in **Fig.** 5, columns 3-6, we examine how batch effects impact gene expression profiles and how these differences vary across cell types.

Fig. 5 shows genomaps constructed using the same visualization procedure for Endothelial (a) and Myeloid (b) cells, enabling assessment of within-tissue similarity and between-tissue contrasts.

The first column of **Fig.** 5a,b shows the original gene expression of the selected cells for each cell type, where distinct expression patterns are observed across the cell types. The second column shows the genomap for the fixed effects reconstruction using the scMEDAL-FE subnetwork, which captures the batch-independent characteristics of the cells. The presence of these distinct gene expression patterns highlights the variability between cell types—independent of batch effects—captured by scMEDAL-FE. In the remaining columns, the following question is explored: If the cells had come from a different batch, how would their gene expression pattern differ? To address this, the cells were projected (reconstructed) as if they had originated from each of four different batches (shown in columns 3-6), including their own batch and three other batches. A red square outline indicates each cell's original (actual) batch. The resulting genomaps demonstrate that scMEDAL-RE subnetwork successfully learned batch-to-batch variability: for the same cell, the reconstructed patterns change across batches in a way that is systematic rather than random. Importantly, specific similarities and contrasts emerge in Fig. 5. For example, septum tissue exhibits consistent patterns (purple squares) for 2 different batches in both Myeloid and Endothelial cells, suggesting strong within-tissue consistency across septum batches. In contrast, septum versus left ventricle projections show markedly different patterns, reflecting robust between-tissue differences.

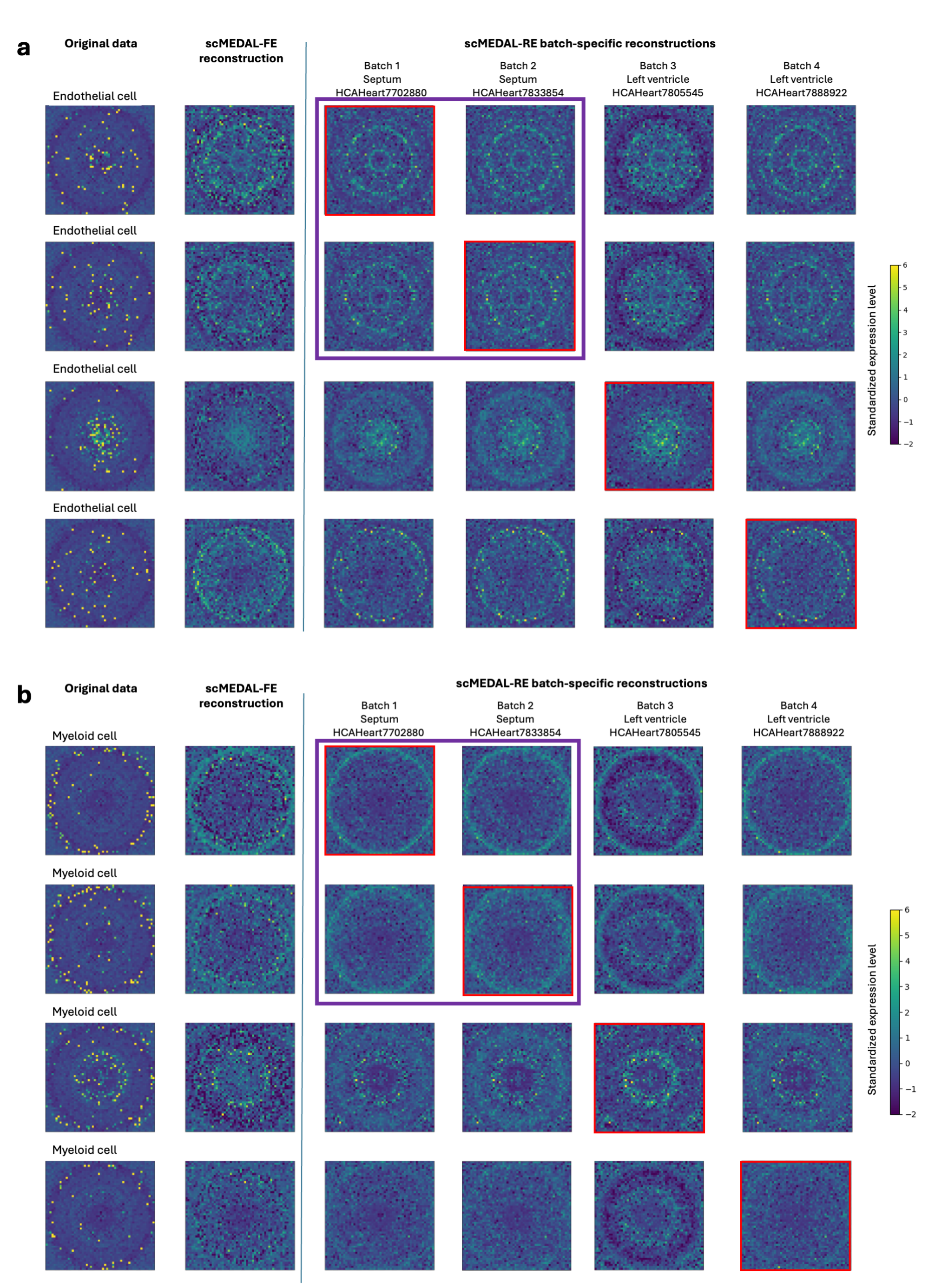

**Fig. 5: How would cells' expression be altered if the cells had originated from different batches in the Healthy Heart dataset?** Genomaps show the gene expression of 2,916 genes for four cells from different cell types across four different batches. The columns show the original expression, the fixed effects reconstruction from scMEDAL-FE, and four counterfactual reconstructions for four different batches using scMEDAL-RE. The red squares highlight the scMEDAL-RE reconstruction for each cell's original batch and the purple squares highlight similarity between two different Septum batches in **a** Endothelial and **b** Myeloid cells.

Taken together, these results demonstrate that scMEDAL framework is capable of capturing and visualizing batch-to-batch variability while preserving biologically meaningful differences between cell types. Unlike traditional methods that obscure such structure, the scMEDAL-RE + consistent genomap combination provides an interpretable, generative framework for probing both shared and divergent features of batch and biological variation.

***ASD Dataset***

In the ASD dataset, the batch effect arises from donor-to-donor variability. The framework captures both donor-specific and donor-agnostic effects. We investigate differences between autistic and healthy individuals by visualizing L2/3 excitatory neurons from three typically developing (control) donors and three donors with a confirmed ASD diagnosis. We select L2/3 excitatory neurons due to their high number of differentially expressed genes between autistic and control subjects[42]. By projecting cells from healthy donors onto the autistic representation—and vice versa—we elucidate gene expression alterations associated with autism. This approach provides insight into disease-specific gene expression patterns while demonstrating the model's capability to capture and visualize diagnostic conditions.

We visualize gene expression profiles of L2/3 cells using genomaps (**Fig.** 6). In the first column, distinct genomap patterns between cells from typically developing donors (**Fig.** 6, rows 1-3) and autistic donors (**Fig.** 6, rows 4-6) highlight autism-associated differences in gene expression. The second column presents the fixed effects reconstruction using the scMEDAL-FE subnetwork, capturing cell-type variability independent of donor effects. To address the question of how a cell's gene expression pattern would appear if it originated from a different donor—specifically, from a subject with autism versus a control—the cells were reconstructed as if they had come from each of several donors, including their own and others (shown in columns three through eight).

We observe that scMEDAL-RE effectively captures donor-specific variability, as evidenced by the varying genomap patterns across each row for the same cell when projected onto different donors. Notably, cells projected onto control and autistic donors exhibit a variety of gene expression patterns, including sets of more actively expressed genes (shown as more pronounced rings), compared to those cells projected onto control donors. The reconstruction corresponding to the cell's original donor is outlined with the red square. Control donors 6032 and 5976 show similar expression profiles in both original and counterfactual projections (purple squares in Fig. 6, third and fourth rows), likely reflecting their similar ages (both from 4-year-old subjects). In contrast, control donor 5242, who was 15 years old, exhibits markedly different expression patterns (red square, first row). All ASD donors display unique profiles consistent with the known heterogeneity of autism. For instance, donor 5419, who experienced sudden unexpected death likely due to epileptic seizure, shows a larger set of genes with elevated expression compared to the other ASD donors (white arrows in Fig. 6, fourth row). These results highlight a central innovation of scMEDAL-RE: its capacity to generate donor-specific counterfactual reconstructions. By predicting how the same cell's expression would appear if it had originated from different donors, scMEDAL-RE not only models batch effects but also provides a mechanistic view of donor heterogeneity and disease-specific variability. This capability goes beyond traditional batch correction by enabling exploration of both shared and individualized transcriptional patterns across diagnostic groups.

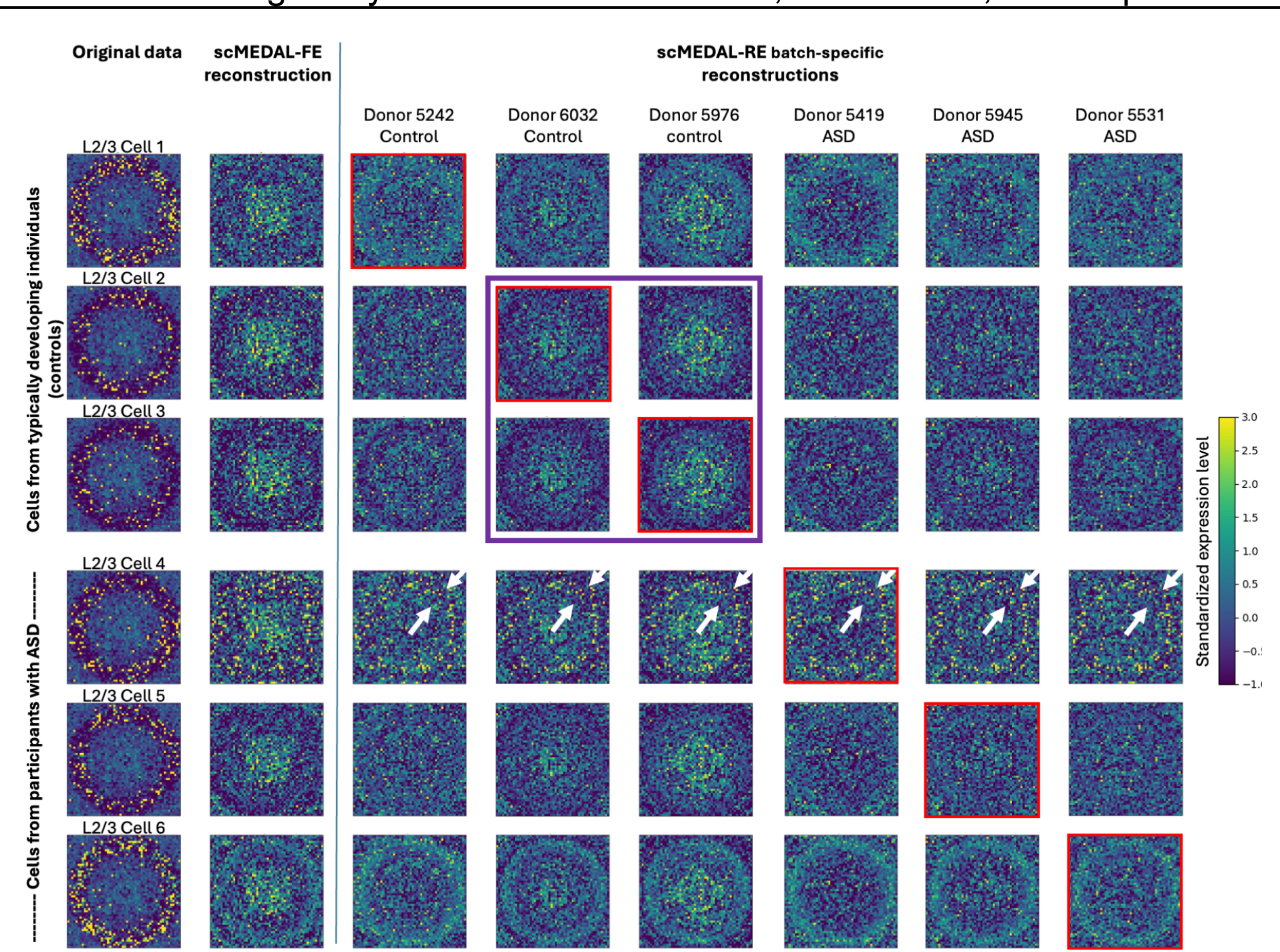

**Fig. 6: How would cells' expression be altered if the cells came from a healthy donor versus a diseased donor in the ASD dataset?** Genomaps show the gene expression of 2,916 genes for six L2/3 cells. The columns depict the cells' original expression, the fixed effects reconstruction from scMEDAL-FE, and six counterfactual reconstructions for six different donors using scMEDAL-RE. The red squares highlight the reconstruction of scMEDAL-RE for each cell's original batch. The purple squares highlight similar genomap patterns of control donors of the same age (4 years old). White arrows highlight ring of high expression pixels in ASD subject 5419.

Using the genomap transform of gene expression values, we further analyzed the scMEDAL-RE projection of 300 L2/3 cells from 15 ASD and 16 control donors. Statistical testing with linear mixed-effects models identified 196 genes significantly associated with ASD compared to controls ($p < 0.05$) (SupplementaryData_1.xls, Supplementary **Table** S13). Notably, **CYFIP1** was downregulated in ASD; given the high dendritic spine density of L2/3 neurons, reduced **CYFIP1** likely disrupts synaptic connectivity and plasticity, consistent with ASD mechanisms[46,47]. **ATP1A3** was higher in ASD (i.e., lower in controls), in line with its role in neuronal excitability and prior reports linking ATP1A3-related disorders to autistic features[48]. **KCNJ2** and **ATP1A2** were overexpressed in ASD and both are associated with epilepsy[49,50], which aligns with the observation that 8 out of 15 ASD patients in the ASD dataset also have epilepsy[42]. **FRYL** was overexpressed in ASD, consistent with previous associations with developmental delay, ASD, and seizures[51] —all conditions represented in our cohort. Finally, a reduced **EOMES**-linked signal was observed in ASD L2/3 neurons, suggesting a developmental footprint consistent with earlier findings on intermediate progenitor cell (IPC) differentiation into L2/3 excitatory neurons and L2/3 depletion under ASD-risk perturbations[52]. Together, these results indicate that genomaps captured donor heterogeneity within the reconstructed spaces of scMEDAL-RE.

### *AML Dataset*

In the leukemia dataset, the batch effect, also referred to as the patient group, arises from the source of the cells, which includes both donors (diseased and controls) and cell lines. To illustrate how scMEDAL-RE captures donor and disease-related variations, we visualize healthy and malignant monocytes from healthy donors, diseased donors, and cell lines. This ability is crucial for understanding gene expression changes associated with malignancy at the individual level and for identifying potential therapeutic targets. It is also important to note that cell lines can have expression patterns that differ fundamentally from those of AML or control donor subjects. This experiment assesses our framework's capacity to project and visualize gene expression profile changes under varying conditions, providing unique insights into biological variability and disease mechanisms that prior batch effect suppression methods do not address.

Both normal cells (**Fig.** 7, rows 2 and 4) and malignant cells (**Fig.** 7, rows 1 and 3) are randomly selected for visualization. The first column shows the original gene expression of both normal and malignant monocytes. We observe that while the overall pattern looks similar, there are numerous variations at the level of individual genes, highlighting the natural variability between donors. The second column presents the fixed effects reconstruction using the scMEDAL-FE subnetwork, which captures cell variability independent of the source (i.e., donor or cell line). In the subsequent columns, the following question is explored: How would a cell's expression pattern vary if it had come from a different donor—specifically from healthy donors, AML donors, or cell lines? To investigate this, both normal and malignant monocytes were reconstructed as if they originated from various donors and cell lines (columns 3-5). We observe that the scMEDAL-RE subnetwork effectively captures donor-specific and disease-specific variability, as evidenced by the varying genomap patterns across each row for the same cell, when projected for each batch. Notably, gene expression patterns differ when normal cells are projected onto AML donors versus healthy donors (e.g., row 2, columns 4-5), and vice versa for malignant cells. These results corroborate our earlier findings in Figure 4: differences between batches become accentuated, and projections for the same AML donor (e.g., AML420B) are more similar to each other than to projections from unrelated donors such as healthy BM5. This demonstrates scMEDAL-RE's ability to model how gene expression would change if a normal or malignant cell had come from a different donor or cell line. The reconstruction corresponding to the original donor or cell line of each cell is indicated by the red square outline. We observe that our framework has learned distinct gene expression patterns between healthy donors, AML donors, and cell lines. For instance, scMEDAL-RE learned variability specific to the MUTZ3 cell line. MUTZ3 cells have adapted to laboratory growth conditions, and this variability differs fundamentally from that of donor samples, explaining the visual differences observed, when normal and malignant cells from donors (rows 2-4) are projected onto the MUTZ3 cell line batch (column 3) compared to cells from the cell line (row 1, column 3). Taken together, these observations underscore the advantage of generative modeling of random effects rather than merely suppressing them, by allowing exploration of how gene expression varies between normal and malignant cells as well as across different donors and batches. Overall, the comprehensive generative scMEDAL-based modeling of both cell-type-specific and donor-specific patterns enhances our understanding of gene expression variability due to disease status and donor differences.

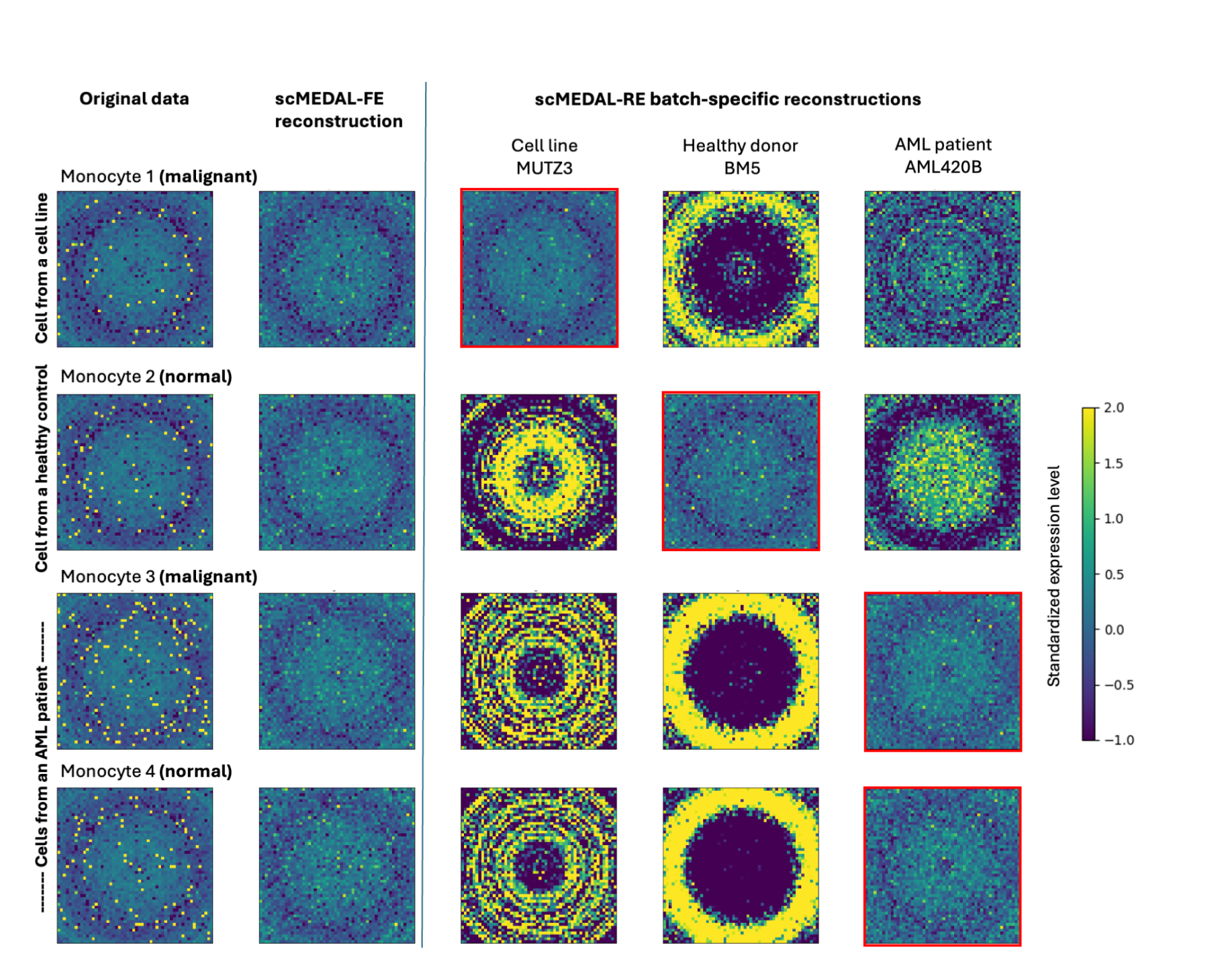

**Fig. 7: How would a cell's expression pattern differ if it had originated from healthy donors, AML donors, or cell lines in the AML dataset?** Genomaps display the gene expression levels of 2,916 genes for normal monocytes (rows 2,4) and malignant monocytes (rows 1,3) across different batches. The illustration includes cells from healthy donors, leukemia patients, and the MUTZ3 cell line. The columns show the original data, the fixed effects reconstruction from scMEDAL-FE, and three counterfactual reconstructions for three different batches using scMEDAL-RE. The red squares highlight the reconstruction of scMEDAL-RE for each cell's original batch.

To further demonstrate this capability, we analyzed expression profiles from 300 monocyte cells for each of the 12 AML (diseased) and 5 control (healthy) **scMEDAL-RE** projections following the genomap transformation. A **Mann–Whitney U test** comparing diseased and healthy projection distributions identified **378 genes** with significant differential expression ($p < 0.05$) (Supplementary Data 1, Supplementary **Table** S14). Several of these genes are closely linked to AML biology. **APC** functions as a tumor suppressor best known in colorectal cancer[53], though rare somatic APC events have also been reported in AML[54]. **AFF3** regulates lymphoid gene programs; when highly expressed—particularly in **HOXA9/MEIS1**-driven[55], lymphoid-leaning stem-like cells—it promotes AML cell self-renewal[56]. **FAT1** normally provides a cellular growth stop signal, but when reduced or mutated (often alongside **FLT3-ITD** and **DNMT3A** mutations), this inhibition is lost[57], leading to unchecked proliferation and poorer clinical outcomes. **PRDM1** (encoding Blimp-1) has been linked to elevated leukemia cell counts in AML patients, suggesting its potential as a biomarker[58]. **FTL**, encoding the ferritin light chain, is frequently upregulated in AML[59], where high ferritin levels are associated with adverse prognosis. **NFATC2** acts as a transcriptional regulator that sustains AML cell proliferation and mitochondrial activity[60]. Additionally, **FPRI** and **SETBP1**, both associated with normal and malignant hematopoietic cell types, were identified, consistent with findings in van Galen et al. (2019)[44]. Collectively, these results highlight molecular alterations relevant to AML pathogenesis and identify potential biomarkers for further investigation.

## 2.6 scMEDAL-RE is complementary to supervised models that enhance cell type preservation

Many transcriptomics datasets include cell-type labels and these can be leveraged by label-informed models to potentially preserve cell-type information in the latent representation while maintaining low reconstruction error. To test this, we benchmarked a standard autoencoder classifier (AEC), scANVI (a semi-supervised variant of scVI), and a label-aware fixed-effects subnetwork (scMEDAL-FEC) that replaces the unsupervised AE in scMEDAL-FE with an autoencoder-classifier for cell type. We then set out to determine whether these label-informed embeddings are complementary to the batch-aware signal captured by the separately trained scMEDAL-RE.

***Healthy Heart dataset***

Introducing a cell type classifier into the base models leads to improved mean ASW scores for both batch effects and cell type within the Healthy Heart dataset (Fig. 8d). The cell-type ASW rose from 0.13 (Fig. 2b) to 0.32 with AEC, a 146% gain; scMEDAL-FEC improved from 0.10 to 0.21 (110% gain). CH and 1/DB clustering scores increased in parallel (Supplementary Table S7 and S9). Although AEC tends to yield slightly higher separability scores than scMEDAL-FEC (differences generally not significant; Fig. 2b, Supplementary Table S9), UMAPs indicate that scMEDAL-FEC better separates cell-type clusters (e.g., Fig 8a ventricular cardiomyocytes, endothelial cells and pericytes) and more effectively reduces protocol separability (Fig. S4). scANVI matches AEC on cell-type ASW (0.32) and improves over scVI (0.22), and it shows the best cross-protocol mixing among the three (Fig. S4). However, scANVI also produces a larger "Not Assigned" cluster (black color) comprising heterogeneous cell types, which can introduce false positive clustering and artificially inflate separability scores. Despite these gains from label supervision, none of the label-informed models separates batch and tissue as effectively as scMEDAL-RE (Figs. 2, 8), underscoring RE's complementary, batch-aware signal, even in supervised

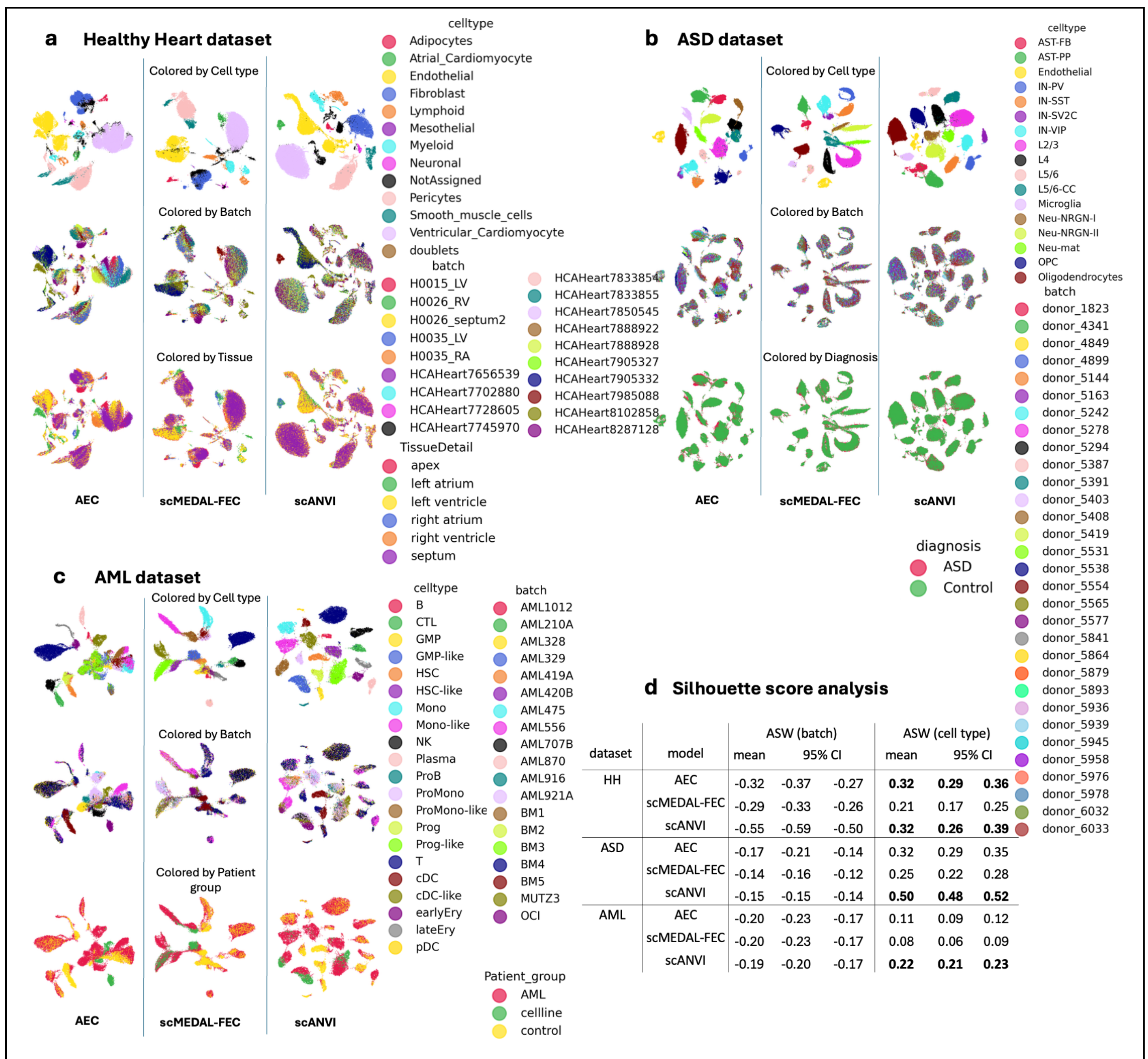

| dataset | model | ASW (batch) | | | ASW (cell type) | | |
|---|---|---|---|---|---|---|---|
| | | mean | 95% CI | | mean | 95% CI | |
| HH | AEC | -0.32 | -0.37 | -0.27 | **0.32** | **0.29** | **0.36** |
| | scMEDAL-FEC | -0.29 | -0.33 | -0.26 | 0.21 | 0.17 | 0.25 |
| | scANVI | -0.55 | -0.59 | -0.50 | **0.32** | **0.26** | **0.39** |
| ASD | AEC | -0.17 | -0.21 | -0.14 | 0.32 | 0.29 | 0.35 |
| | scMEDAL-FEC | -0.14 | -0.16 | -0.12 | 0.25 | 0.22 | 0.28 |
| | scANVI | -0.15 | -0.15 | -0.14 | **0.50** | **0.48** | **0.52** |
| AML | AEC | -0.20 | -0.23 | -0.17 | 0.11 | 0.09 | 0.12 |
| | scMEDAL-FEC | -0.20 | -0.23 | -0.17 | 0.08 | 0.06 | 0.09 |
| | scANVI | -0.19 | -0.20 | -0.17 | **0.22** | **0.21** | **0.23** |

**Fig. 8. Using supervised training improves the preservation of cell-type information in AEC, scMEDAL-FEC, and scANVI but does not decouple biological variation from batch effects (tissue, diagnosis, patient group) like scMEDAL-RE. a-c** UMAP visualizations of gene expression in the latent spaces across different datasets from the AEC, scMEDAL-FEC and scANVI models. **a** UMAP visualization from 41544 cells across 20 batches of the Healthy Heart dataset. **b** UMAP visualization computed from 62,735 cells in the ASD dataset. **c** UMAP visualization computed from 23,049 cells in the AML dataset. **d** Avg. Silhouette Width (ASW) scores (mean & 95% confidence intervals across 5 folds) for batch and cell-type separability in the latent spaces of Healthy Heart, ASD, and AML datasets, comparing AEC, scMEDAL-FEC and scANVI. Bolded scores indicate the highest separability for cell-type ASW.

scenarios.

### *ASD dataset*

For the ASD dataset, adding a classifier boosts cell-type separability: AEC increases cell-type ASW from 0.20 (AE) to 0.32 (+60%), scMEDAL-FEC from 0.17 to 0.25 (+47%), and scANVI reaches 0.50 versus 0.34 for scVI (+47%), with concordant improvements in 1/DB and CH (Supplementary Tables S7, S10). Even so, scMEDAL-RE provides the clearest separation of batch and diagnosis (Figs. 2, 8), highlighting its role as a complementary add-on that restores donor/diagnosis structure often blurred by label-informed embeddings.

***AML dataset***

For the leukemia dataset, classifier supervision similarly improves cell-type ASW, albeit from lower baselines: AEC rises from 0.06 (AE) to 0.11 (+83%), scMEDAL-FEC from 0.05 to 0.08 (+60%), and scANVI from 0.03 (scVI) to 0.22 (+633%), with consistent CH and 1/DB gains (Tables S8, S11). Nevertheless, scMEDAL-RE remains superior for batch/patient-group separation (Figs. 2, 8).

## 2.7 Improving cell classification by leveraging scMEDAL-RE across diverse embeddings

The scMEDAL's random effects component (scMEDAL-RE) models batch effects quantitatively, rather than discarding them, reversing the usual paradigm in which batch structure is treated as nuisance signal to be removed. This improves the framework's ability to predict cell properties, like cell type, patient group, and disease diagnosis.

We evaluated the performance predicting such properties across three datasets. For each embedding method, we trained a Random Forest classifier twice: first using the method's latent space alone (e.g., PCA; scMEDAL-FE; scVI; Harmony; Scanorama; SAUCIE; AEC; scMEDAL-FEC; scANVI), and then using the same latent space concatenated with the scMEDAL-RE projection (e.g., PCA+scMEDAL-RE, scMEDAL-FE+scMEDAL-RE, scVI+scMEDAL-RE, etc.). This design isolates the value of adding scMEDAL-RE to each baseline embedding. Because the RE projection contributes signal that is complementary to these baselines, we hypothesize that it can be productively paired with diverse batch-correction pipelines to improve downstream prediction. Prior to training, all latent spaces were standardized, adjusting each to have a mean of zero and a standard deviation of one, to ensure comparability of the inputs.

Regarding prediction targets, we focused on outcomes that carry biological signal intertwined with batch. In the Healthy Heart dataset, the primary target was tissue of origin within the heart. For the ASD dataset, we used disease diagnosis (autism spectrum disorder or typically developing) as the Dx target. In the AML dataset, the primary target was patient group—which includes diseased (AML patients), healthy donors, and cell line classes—to assess the model's capability to differentiate among these groups. Since all three datasets[41,42,44] included cell type annotations, we also evaluated classification by cell type for unsupervised models. For AML specifically, malignant and healthy cells from the same cell type were treated as two distinct cell-type labels. To ensure consistency in our evaluation, we employed the same 5-fold cross-validation splits used to obtain scMEDAL latent spaces for all classification tasks.

In Table 1, augmenting each method's latent space with scMEDAL-RE increases balanced accuracy for all primary targets—tissue (Healthy Heart), diagnosis (ASD), and patient group (AML)—across all datasets and models, relative to using the embedding alone. For Healthy Heart (tissue-of-origin), it exceeds 97% with scMEDAL-RE, versus 31.03% for Harmony and 58.34% for PCA alone. For ASD (diagnosis), balanced accuracy exceeds 94% with scMEDAL-RE for every model, compared with 57.06% for Harmony and 73.00% for PCA when used alone. For AML (patient group), it surpasses 99.8% with scMEDAL-RE, versus 38.27% for Harmony and 89.19% for PCA alone. Across standalone embeddings, Harmony consistently yields the lowest balanced accuracy and PCA the highest in each dataset. The relatively strong PCA baselines are expected: because PCA does not correct batch effects, it retains biological signal linked to tissue, donor, and patient group The benefits from scMEDAL-RE reflect the results of Figures 2–4, where scMEDAL-RE cleanly resolves tissue, diagnosis, and patient-group structure.

For the cell-type targets (Supplementary Table S12), improvements were smaller overall. In Healthy Heart, Harmony increased from 73.72% to 81.88% balanced accuracy when combined with scMEDAL-RE; most other methods also improved, whereas PCA showed no gain. This pattern likely reflects that several batch-correction methods tend to merge atrial and ventricular cardiomyocytes (Fig. 2), limiting separability. In ASD, all models without scMEDAL-RE already exceeded 88% balanced accuracy, making the task easier and yielding only mild gains with scMEDAL-RE for most methods (again no improvement for PCA). We also note that many batch-correction methods merged AST-PP and AST-FB classes (Fig. 3), whereas PCA preserved their separation. In AML, Harmony improved from 49.24% to 71.42% with scMEDAL-RE, and other methods showed modest gains. This is consistent with AML being the most challenging dataset for cell-type discrimination in the latent spaces (Fig. 4).

**Table 1. Combining various embeddings with batch-specific (scMEDAL-RE) latent spaces improves the accuracy of cellular level predictions for targets that contain biological information.** Mean results across 5 folds and 95% confidence intervals (CI) for a Random Forest Classifier using tissue (Healthy Heart), diagnosis (ASD), and patient group (AML) as targets. The performance metrics include accuracy, balanced accuracy, and chance accuracy. Boldface denotes cases where pairing a model with scMEDAL-RE yields higher balanced accuracy, and the combined model also surpasses the standalone model in overall accuracy.

| | | | Accuracy | | | Balanced Accuracy | | | Chance Accuracy | | |
|---|---|---|---|---|---|---|---|---|---|---|---|
| **Dataset** | **Target** | **Latent space** | **mean** | **95%CI** | | **mean** | **95%CI** | | **mean** | **95%CI** | |
| **HH n=486,134** | Tissue (k=6) | Harmony | 33.25 | 33.05 | 33.46 | 31.03 | 30.72 | 31.33 | 17.91 | 17.73 | 18.08 |
| | | Harmony+scMEDAL-RE | 97.85 | 96.39 | 99.31 | **97.76** | **96.4** | **99.11** | | | |
| | | PCA | 57.97 | 57.77 | 58.16 | 58.34 | 58.13 | 58.55 | | | |
| | | PCA+scMEDAL-RE | 97.84 | 96.54 | 99.14 | **97.72** | **96.51** | **98.94** | | | |
| | | SAUCIE | 50.31 | 49.31 | 51.31 | 51.28 | 50.26 | 52.29 | | | |
| | | SAUCIE+scMEDAL-RE | 97.42 | 95.88 | 98.97 | **97.3** | **95.89** | **98.7** | | | |
| | | Scanorama | 51.91 | 51.51 | 52.3 | 51.74 | 51.2 | 52.27 | | | |
| | | Scanorama+scMEDAL-RE | 98.14 | 97.04 | 99.24 | **98.04** | **97.01** | **99.07** | | | |
| | | scANVI | 35.73 | 35.11 | 36.34 | 34.94 | 34.27 | 35.62 | | | |
| | | scANVI+scMEDAL-RE | 97.44 | 95.85 | 99.02 | **97.31** | **95.82** | **98.81** | | | |
| | | scMEDAL-FE | 40.21 | 39.52 | 40.89 | 39.63 | 38.87 | 40.38 | | | |
| | | scMEDAL-FE+scMEDAL-RE | 97.66 | 96.06 | 99.26 | **97.56** | **96.09** | **99.04** | | | |
| | | scMEDAL-FEC | 45.36 | 44.66 | 46.06 | 45.29 | 44.62 | 45.96 | | | |
| | | scMEDAL-FEC+scMEDAL-RE | 97.54 | 96.03 | 99.05 | **97.43** | **96.03** | **98.83** | | | |
| | | scVI | 40.4 | 39.67 | 41.14 | 40.75 | 39.86 | 41.64 | | | |
| | | scVI+scMEDAL-RE | 97.72 | 96.3 | 99.14 | **97.62** | **96.3** | **98.93** | | | |
| **ASD n=104,559** | Diagnosis (k=2) | Harmony | 57.05 | 56.61 | 57.48 | 57.06 | 56.62 | 57.49 | 50.06 | 49.75 | 50.37 |
| | | Harmony+scMEDAL-RE | 95.11 | 93.19 | 97.03 | **95.12** | **93.2** | **97.03** | | | |
| | | PCA | 72.99 | 72.59 | 73.39 | 73 | 72.6 | 73.4 | | | |
| | | PCA+scMEDAL-RE | 94.7 | 92.64 | 96.75 | **94.7** | **92.65** | **96.76** | | | |
| | | SAUCIE | 70.57 | 69.44 | 71.7 | 70.58 | 69.44 | 71.72 | | | |
| | | SAUCIE+scMEDAL-RE | 95.05 | 92.9 | 97.19 | **95.06** | **92.91** | **97.2** | | | |
| | | Scanorama | 65.75 | 64.19 | 67.31 | 65.76 | 64.2 | 67.32 | | | |
| | | Scanorama+scMEDAL-RE | 95.27 | 93.38 | 97.17 | **95.28** | **93.39** | **97.17** | | | |
| | | scANVI | 59.88 | 59.7 | 60.06 | 59.88 | 59.7 | 60.06 | | | |
| | | scANVI+scMEDAL-RE | 95.14 | 93.12 | 97.17 | **95.15** | **93.13** | **97.17** | | | |
| | | scMEDAL-FE | 67.23 | 66.65 | 67.82 | 67.24 | 66.65 | 67.83 | | | |
| | | scMEDAL-FE+scMEDAL-RE | 94.87 | 92.82 | 96.92 | **94.88** | **92.83** | **96.93** | | | |
| | | scMEDAL-FEC | 62.1 | 61.21 | 62.98 | 62.11 | 61.23 | 62.99 | | | |
| | | scMEDAL-FEC+scMEDAL-RE | 95.18 | 93.27 | 97.08 | **95.18** | **93.28** | **97.09** | | | |
| | | scVI | 63.14 | 62.71 | 63.56 | 63.13 | 62.7 | 63.56 | | | |
| | | scVI+scMEDAL-RE | 95.32 | 93.4 | 97.24 | **95.33** | **93.41** | **97.25** | | | |
| **AML n=38,417** | Patient group (k=3) | Harmony | 76.5 | 76.1 | 76.9 | 38.27 | 36.74 | 39.8 | 60.82 | 60.37 | 61.27 |
| | | Harmony+scMEDAL-RE | 99.96 | 99.91 | 100.01 | **99.92** | **99.83** | **100.01** | | | |
| | | PCA | 94.04 | 93.18 | 94.9 | 89.19 | 87.17 | 91.21 | | | |
| | | PCA+scMEDAL-RE | 99.95 | 99.9 | 100.01 | **99.91** | **99.81** | **100.01** | | | |
| | | SAUCIE | 91.8 | 90.67 | 92.93 | 86.21 | 83.75 | 88.66 | | | |
| | | SAUCIE+scMEDAL-RE | 99.94 | 99.88 | 100 | **99.89** | **99.79** | **100** | | | |
| | | Scanorama | 85.6 | 84.44 | 86.76 | 67.36 | 63.25 | 71.46 | | | |
| | | Scanorama+scMEDAL-RE | 99.95 | 99.9 | 100 | **99.91** | **99.81** | **100.01** | | | |
| | | scANVI | 89.87 | 89.53 | 90.21 | 78.09 | 76.58 | 79.61 | | | |
| | | scANVI+scMEDAL-RE | 99.95 | 99.9 | 100 | **99.91** | **99.82** | **100** | | | |
| | | scMEDAL-FE | 91.56 | 90.96 | 92.15 | 82.97 | 81.9 | 84.04 | | | |
| | | scMEDAL-FE+scMEDAL-RE | 99.95 | 99.9 | 100 | **99.91** | **99.82** | **99.99** | | | |
| | | scMEDAL-FEC | 82.12 | 78.74 | 85.5 | 56.92 | 48.25 | 65.58 | | | |
| | | scMEDAL-FEC+scMEDAL-RE | 99.95 | 99.9 | 100 | **99.9** | **99.81** | **99.99** | | | |
| | | scVI | 88.43 | 87.61 | 89.24 | 76.13 | 72.9 | 79.37 | | | |
| | | scVI+scMEDAL-RE | 99.94 | 99.88 | 99.99 | **99.88** | **99.78** | **99.98** | | | |

Collectively, these results demonstrate that the scMEDAL framework is plug-and-play compatible with existing batch correction methods. Because the scMEDAL-RE component is trained separately from scMEDAL-FE, the RE component and accompanying visualizations of scMEDAL framework are easily combined with existing methods, affording consistent improvement to cell classification by supplying donor-aware information that existing embeddings either remove or fail to capture

# 3. Discussion

scMEDAL is a mixed-effects deep learning framework that separates batch-invariant and batch-specific variation via modular fixed-effects (FE) and random-effects (RE) subnetworks. The primary innovation is the scMEDAL-RE network and associated visualization components. Instead of removing batch information, this network learns batch distributions quantitatively, treating tissue, diagnosis, and patient group (donor, cell-line, control) differences as biologically meaningful signal, rather than nuisance. This not only gives give insight into the data acquisition process, but makes batch structure analyzable, comparable across conditions, and directly useful for downstream prediction and visualization.

Unlike other batch correction methods (e.g., AutoClass[24], scVI[14], Harmony[8], Scanorama[9], SAUCIE[27], scANVI[15], iMAP[17], ResPAN[18], scDREAMER[19], IMAAE[20], ABC[21], DB-AAE[22]) which learn batch-invariant embeddings without explicitly modeling batch distributions, scMEDAL-RE models these distributions in its RE subnetwork. We applied scMEDAL across three use cases—tissue of origin (Healthy Heart), diagnosis (ASD vs. control), and patient group in AML—where the “batch” is heterogeneous (donor, cell line, control) which contain important biological distinctions that should not be collapsed into a single category.

Our first contribution is to demonstrate that scMEDAL-RE visualizations recover clear, biologically meaningful structure that self-supervised batch-invariant embeddings, such as scVI, Harmony, Scanorama and SAUCIE, and traditional dimensionality-reduction methods —including PCA and AE, tend to mask. UMAPs computed directly from the scMEDAL-RE embedding yield batch-aware maps that reveal the structure of batch heterogeneity.

Our second contribution concerns the scenario when cell-type labels are available. Here we show that scMEDAL-RE is complementary to supervised models that use cell-type labels during training to enhance cell-type biology in the latent space. We incorporated a cell-type classifier into scMEDAL-FE to maintain cell-type signals within the batch-invariant space. Methods like AutoClass[24], scDREAMER[19], and ABC[21] and scANVI[15] use cell-type labels during training for batch effect suppression and preservation of cell-type structure. However, low-quality or uncertain labels can lead to spurious clustering and incorrect interpretations, as we observed in the scANVI model enhancing “Not Assigned” cells clustering (Section 2.6), but which is mitigated through our supervised subnetwork, scMEDAL-FEC.

Our third contribution is to enable retrospective "What if?" analyses by leveraging the generative nature of the framework’s default fixed and random effects subnetworks to predict a cell's expression under different batches or diagnoses, in addition to the batch agnostic expression. Using the learned, modeled random effects, we reconstructed a cell's gene expression under different conditions to answer “What if?” questions. For example, we projected how a cell's profile would change if it originated from a different batch, highlighting variability within and between cell types. In the ASD and AML datasets, we simulated how cells would appear from donors with different diagnoses, providing insights into disease and donor variability. Our model's ability to project cells across batches represents a significant advancement in personalized single-cell modeling. Models like AIF[25] incorporate batch modeling but do not explicitly separate batch effects. By isolating batch variability, our approach enhances explainability and facilitates the disentanglement of batch effects. The scMEDAL framework is the first to exploit cell projections for addressing such retrospective scenarios (“What if?” questions) and to separate fixed and random effects in scRNA-seq data using deep learning.

Historically, evaluation of batch correction methods has focused on clustering metrics[4,5,61,62] rather than classification performance. However, we hypothesized that predictive models, which learn complementary batch-agnostic and batch-specific latent spaces, would enable more accurate cell-level predictions. Our fourth contribution is to establish that scMEDAL-RE improves classification when existing batch correction methods are used interchangeably with the scMEDAL framework. The framework including scMEDAL-RE network and all of the visualization components (consistent genomaps, UMAPs, and multi-metric evaluation) can be used with any batch corrected embedding, such as our scMEDAL-FE, traditional PCA or AE, as well as unsupervised models (scVI, Harmony, Scanorama, SAUCIE), and label-informed AEC, scANVI and scMEDAL-FEC. We observe the largest gains for targets where biological signal is intertwined with batch. In all compared base methods—including scMEDAL-FE, PCA, scVI[14], Harmony[8], Scanorama[9], SAUCIE[27], we observed scMEDAL-RE adds complementary, non-redundant information and improves cell-level classification accuracy without requiring any retraining of base models..

One potential limitation of this work is the use of hyperparameter optimization, such as for the weights assigned to loss function terms for the subnetworks; however, in practice we have found that even standard strategies such as grid search suffice to readily find weights that balance the loss terms well.

The scMEDAL framework is modular, allowing for independent training of both the fixed (batch effect suppression) and Bayesian random effects subnetworks. This plug-and-play design lets users swap in their preferred batch-suppression method while retaining scMEDAL's overall architecture, the random-effects module (scMEDAL-RE), and its visualization tools. We demonstrate this by replacing the fixed-effects component with scVI[14], Harmony[8], Scanorama[9], SAUCIE[27], scANVI[15] and we anticipate this result generalize to batch-correction methods that do not explicitly model batch effects. The above contributions, together with the comprehensive framework evaluation across diverse single cell and single nucleus RNA-seq datasets spanning multiple health conditions, cell-types, and batch effects—including datasets from the cardiovascular system (Healthy Heart), ASD, and AML, underscore the value of separately modeling and quantifying fixed and random (batch) effects through generative deep learning for interpretable transcriptomics analysis.

# 4. Methods

## 4.1 Mathematical foundation of the scMEDAL framework

When data is confounded by batch effects, as commonly observed in single-cell RNA sequencing (scRNA-seq) data, it is essential to employ methods that account for the confounding, otherwise cell types can become mislabeled and transcriptomic impacts of diseases can be misconstrued. The scMEDAL framework addresses this need to mitigate batch confounding. The framework ), is inspired by the mathematical foundation of statistical mixed effects models (LME) model [63]. The LME model is defined as:

$$y_i = \beta_0 + \boldsymbol{x}_i^T \boldsymbol{\beta}_1 + u_{j,0} + \boldsymbol{x}_i^T \boldsymbol{u}_j + \epsilon_i \qquad (1)$$

where $y_i$ is the target prediction for the $i^{th}$ sample which is a cell in our application, $x_i^T$ is the observed gene expression vector of the $i^{th}$ cell, while $\beta_0$ and $\boldsymbol{\beta}_1$ denote the fixed effects scalar intercept and slope vector, respectively, which capture batch invariant trends across all samples. Batches (technical and biological) cluster the cells' expression, causing greater correlation among cells within the same batch than between batches. In the mixed effects model, the $i^{th}$ cell comes from the $j^{th}$ batch. The random effects intercept and slope vector are denoted $u_0$ and $\boldsymbol{u}_{j,0}$ respectively, and are assumed to follow a random distribution, most often modeled as a multivariate normal distribution with mean 0 i.e., $\boldsymbol{u} \sim N(0, \boldsymbol{\Sigma}_j)$. The residual error term of the $i^{th}$ sample is denoted $\epsilon_i$. In traditional LME models, fixed effects are estimated jointly with random effects: the fixed effects are obtained via the Best Linear Unbiased Estimator (BLUE), while the random effects are predicted via the Best Linear Unbiased Predictor (BLUP), with both components mutually dependent through maximum likelihood optimization. In contrast, scMEDAL trains the fixed effects (FE) and random effects (RE) subnetworks separately, and they only interact downstream when their embeddings are combined for classification. This design choice enhances scalability and efficiency in high-dimensional single-cell data, though it departs from the joint estimation principle of LME. Thus, while scMEDAL is not "built upon" LME in the formal statistical sense, it remains conceptually inspired by the idea of decomposing variation into interpretable fixed (biological signal) and random (batch-associated) components.

The proposed scMEDAL framework implements all the components of the mixed effects foundation using neural networks, resulting in a novel model capable of learning both linear and nonlinear relationships between $\boldsymbol{x}_i^T$ and $y_i$. Additionally, it mitigates the confounding batch (clustering) effects by separately quantifying both fixed and random effects, enhancing model interpretability, data understanding, and cell-level predictions.

## 4.2 The Bayesian neural network architecture of the scMEDAL model

The following sections present a description of the scMEDAL framework components, which quantitatively and separately model both fixed and random effects within gene expression vectors of individual cells. The framework's architecture is comprised of two parallel autoencoder subnetworks. The *fixed effects subnetwork* includes the first autoencoder, a conventional autoencoder neural network (**Fig.** 1, blue area) and an adversarial classifier $A$, (**Fig.** 1, gray area) and together they learn the batch-invariant features. Our primary innovation is the random-effects subnetwork, scMEDAL-RE : (Fig. 1, orange area) a Bayesian autoencoder with probabilistic weights that learns batch-dependent structure without collapsing biological signal. The FE and scMEDAL-RE subnetworks are trained independently, and their embeddings are combined by a learned mixing function $f_M$ (denoted mixed effects classifier) that combines the fixed and random effects for prediction. The weights within scMEDAL are learned from the training data partition, which includes a gene expression count matrix and a batch design matrix. The gene expression count matrix is $\boldsymbol{X} \in \mathbb{R}^{n \times g}$, where $n$ is the number of cells in the training partition, and $g$ is the number of genes. The $i^{th}$ row of $\boldsymbol{X}$ contains the gene expression, $\boldsymbol{x}_i$, of the $i^{th}$ cell. To encode batch membership information for the $n$ cells and $K$ batches, we introduce a one-hot encoded design matrix $\boldsymbol{Z} \in \mathbb{R}^{n \times K}$, where $\boldsymbol{Z}_{i,j} = 1$ if sample $i$ belongs to cluster $j$ and $\boldsymbol{Z}_{i,j} = 0$ otherwise. The one hot encoding of the $i^{th}$ cell is the $i^{th}$ row, $\boldsymbol{z}_i$, from matrix $\boldsymbol{Z}$. Each subnetwork learns a lower dimensional latent representation of $\boldsymbol{x}_i$ and outputs a gene expression $\widehat{\boldsymbol{x}}_i$ reconstructed from its latent space representation.

### 4.2.1 Fixed Effects Subnetwork

We start by describing the architecture of the conventional autoencoder (**Fig.** 1, blue area), which contains an encoder to compress a gene expression vector $\boldsymbol{x}$ into a latent, **F**ixed effects representation $e_F(\boldsymbol{x}; \boldsymbol{\beta})$, where $\boldsymbol{\beta}$ contains all learned weights up to and including the latent representation. A decoder then reconstructs the gene expression vector $\hat{\boldsymbol{x}}_F$. The encoder comprises three blocks, each containing two dense layers. The output of the final encoder block is passed through a dense layer with two neurons, generating the compressed latent representation. All layers, except the decoder's output layer, use the Scaled Exponential Linear Unit (SELU) activation function. The decoder follows an architecture symmetric to the encoder, also consisting of three layers, with a linear activation applied in the output dense layer. For the supervised fixed effects subnetwork *variant*, we assume cell type information is available for training and test its efficacy. In this scenario, to simultaneously perform cell type classification, we introduce an auxiliary classifier subnetwork (**Fig.** 1, middle of blue area) which predicts the cell type, $\hat{y}_F$, from the latent representation. This auxiliary classifier takes the encoder's latent representation as input and contains a dense hidden layer and a softmax output layer.

To create the fixed effects subnetwork, which is a domain adversarial autoencoder that we denote as scMEDAL-FE, we add an adversarial classifier $A$, (**Fig**. 1, gray area), to predict each cell's batch $\hat{\boldsymbol{z}}_A$ from the layer activations of the encoder. To the extent that it can predict the batch (i.e., $\hat{\boldsymbol{z}}_A = \boldsymbol{z}$) it penalizes the FE autoencoder for using batch-dependent features. Thus through a generalization loss, the autoencoder is penalized for learning features that allow accurate batch prediction. Note that the encoder and decoder weights are tied so that this penalty affects both modules of the autoencoder. The adversary uses the same architecture as the encoder, with the addition of a final softmax output layer. The overall loss function $\mathcal{L}_{FE}$ of the fixed effects subnetwork consists of this categorical cross entropy penalty term and a reconstruction error term which is expressed as:

$$\mathcal{L}_{FE}(\boldsymbol{X}, \boldsymbol{Z}) = \lambda_{recon,F} \mathcal{L}_{MSE}\left(\boldsymbol{X}, \widehat{\boldsymbol{X}}\right) - \lambda_A \mathcal{L}_{CCE}\left(\boldsymbol{Z}, \widehat{\boldsymbol{Z}}\right) \qquad (2)$$

where the hyperparameters $\lambda_{MSE}$ and $\lambda_A$ control the relative weight of the reconstruction and generalization loss, respectively. The reconstruction error is measured as the mean squared error (MSE):

$$\mathcal{L}_{MSE}\left(\boldsymbol{X}, \widehat{\boldsymbol{X}}\right) = \frac{1}{n} \sum_{i=1}^{n} \left|\left|\boldsymbol{x}_i - \widehat{\boldsymbol{x}}_i\right|\right|^2 \qquad (3)$$

where $\boldsymbol{x}_i$ and $\hat{\boldsymbol{x}}_i$ are the original and reconstructed gene expression vectors of cell $i$. The categorical cross-entropy (CCE) quantifies the adversarial classifier's performance:

$$\mathcal{L}_{CCE}\left(\boldsymbol{Z},\hat{\boldsymbol{Z}}\right) = -\frac{1}{n}\sum_{i=1}^{n}\sum_{k=1}^{K}\boldsymbol{z}_{i,k}\log\hat{\boldsymbol{z}}_{i,k} + \left(1-\boldsymbol{z}_{i,k}\right)\log\left(1-\hat{\boldsymbol{z}}_{i,k}\right) \quad (4)$$

where $\boldsymbol{z}_{i,k}$ is the true label (one-hot encoded) and $\hat{\boldsymbol{z}}_{i,k}$ is the predicted probability for class $k$ for cell $i$. Minimizing the fixed effects loss, $\mathcal{L}_{FE}$, enables this subnetwork to suppress batch-specific variations and capture batch-agnostic fixed effects.

scMEDAL-FE is a domain-adversarial autoencoder, not a GAN: its training objective and loss differ from scGAN[16] and other GAN-based autoencoders. In practice, scMEDAL-FE is our default fixed-effects module for the proposed scMEDAL framework, but it can be swapped for any method that does not explicitly model batch effects (e.g., scGAN[16], scVI, scVI[14], Harmony[8], Scanorama[9], SAUCIE[27], scANVI[15] ) without changing the overall scMEDAL framework, including the visualization approaches and the scMEDAL-RE subnetwork. Unlike scVI, which explicitly conditions both its inference and generative networks on the batch covariate by concatenating batch to the inputs and latent to discourage redundant batch encoding, scMEDAL-FE does not receive batch labels as input; instead, it uses domain-adversarial training to actively purge batch signal from the latent representation. Thus, while scVI and scMEDAL-FE share the goal of batch-invariant embeddings, they achieve it via different mechanisms. By contrast, our key innovation, the random-effects subnetwork scMEDAL-RE, is designed to capture batch: it conditions on batch and, through variational inference and reconstruction, learns cell-specific, batch-conditional deviations rather than suppressing them.

### 4.2.2 Random Effects Bayesian Subnetwork

The random effects subnetwork (scMEDAL-RE) specifically models the batch-dependent effects. Because gene expression batch effects are believed to pervade all levels of the feature hierarchy, including lower gene-level features as well as higher-level gene abstraction features, we construct the random effects subnetwork (**Fig.** 1, orange area), as a second, autoencoder whose depth mirrors the FE conventional autoencoder. However, this second autoencoder, is a *Bayesian* network which we denote as scMEDAL-RE, and thus it has probabilistic weights. This Bayesian autoencoder consists of dense hidden layers in its encoder and decoder components (**Fig.** 1, left side of orange box). The input consists of the standardized expression matrix, $\boldsymbol{X}$, for cells in the training set, and the label matrix for the cells indicating their batch, $\boldsymbol{Z}$, with additional connections providing $\boldsymbol{Z}$ directly to each layer. The autoencoder learns a latent representation which can be used to faithfully reconstruct the input as its output. The RE subnetwork also includes a batch classifier network (**Fig.** 1, right side of orange box) which predicts the batch label $\boldsymbol{z}$ for each cell using the latent representation of the RE autoencoder. To the extent that it can predict the batch (i.e., $\hat{\boldsymbol{z}}_L = \boldsymbol{z}$) it rewards the RE autoencoder weights that construct that latent representation.

The Bayesian random effects subnetwork is a Bayesian autoencoder, $\hat{\boldsymbol{x}}_R = h_R\left(\boldsymbol{x}; U(\boldsymbol{z})\right)$, which encodes a given cell's gene expression, $\boldsymbol{x}$, into a lower dimensional, batch effect-laden, latent representation, $e_R(\boldsymbol{x}; U(\boldsymbol{z}))$, and then decodes that representation back to a reconstruction, $\hat{\boldsymbol{x}}_R$, which closely approximates $\boldsymbol{x}$. Collectively we represent the probabilistic weights of all the layers the autoencoder as $U(\boldsymbol{z})$ and we note that these weights in general, depend on the batch from which the cell came, $\boldsymbol{z}$. The Bayesian approach entails finding these weights, $U$, that are most likely given the observed genomic data $\boldsymbol{X}$. We do this by inferring the posterior distribution $p(U \mid X)$ ), for all the cells in the training set, $X$. In scMEDAL, it is achieved through variational inference (VI), which casts the probabilistic Bayesian modeling as an optimization problem that can be solved efficiently through gradient descent.[64] In VI, the intractable posterior $p(U \mid \boldsymbol{X})$, is approximated with a surrogate $q(U)$, chosen from a tractable family (multivariate normal), by maximizing the ELBO which is also minimizing $D_{\mathrm{KL}}(q(U) \parallel p(U \mid X))$. Closeness between two distributions can be measured with the Kullback-Leibler (KL) divergence:

$$D_{\mathrm{KL}}(q(U) \parallel p(U \mid \boldsymbol{X})) = \mathbb{E}_{q(U)}\left[\log \frac{q(U)}{p(U|\boldsymbol{X})}\right] \quad (5)$$

Using Bayes rule we can expand the posterior in the expectation, $p(U \mid \boldsymbol{X}) = \frac{p(X|U)p(U)}{p(\boldsymbol{X})}$, however we cannot compute all its factors, because of the intractable marginalization $p(\boldsymbol{X})$ which requires integrating over all possible values of $U$. VI addresses this issue by maximizing the Evidence Lower Bound (ELBO), which provides an approximation to the true posterior $p(\ U \mid \boldsymbol{X}\ )$. After expansion and regrouping terms, the ELBO constraint is derived and it identifies the surrogate posterior which maximizes the log-likelihood of the data, and contains entirely tractable terms:

$$\mathrm{ELBO} = \mathbb{E}_q[\log\ p(\boldsymbol{X} \mid U)] - D_{\mathrm{KL}}(q(U) \parallel p(U)) \quad (6)$$

where the first right-hand term is the expected log-likelihood of the gene expression data and the second term is the KL divergence between the surrogate posterior and the prior over model weights.[64] The scMEDAL framework is a deep learning neural network, which is optimized via gradient *descent*. To walk downhill, we minimize the negative ELBO. The first term is the data fidelity term, $E_{q(U)}[log\ p(\boldsymbol{X}|U)]$, which measures how well the model fits the data, which we measure as the reconstruction error, $\mathcal{L}_{MSE}(\boldsymbol{X}, \widehat{\boldsymbol{X}})$. After substitution, the objective loss function for the RE subnetwork is:

$$\mathcal{L}_{recon,R}(\boldsymbol{X}, \widehat{\boldsymbol{X}}) + \lambda_K D_{\mathrm{KL}}(q(U) \parallel p(U)) \quad (7)$$

where $\lambda_K$ is a hyperparameter that controls the degree of regularization from the second term.

The random effect weights of scMEDAL-RE are learned layer-wise using Bayesian, random effects dense (REDEN) blocks (Supplementary **Fig.** S1). REDEN blocks supplant the layers of the original autoencoder. They begin with a dense layer, which takes as input the learned representation output from the previous layer $\boldsymbol{x}_{l-1}$ and output the learned representation $\boldsymbol{d}_l$. Then a learned batch-specific random slope, $\gamma(\boldsymbol{z})$, and batch-specific bias, $\mathrm{b}(\boldsymbol{z})$, is applied to $\boldsymbol{d}_l$, effectively rescaling and shifting the dense layer outputs in a batch-dependent manner, into a new transformed learned representation, $\boldsymbol{x}_l$. These batch-specific slopes $\gamma(\boldsymbol{z}) \sim N(0, \sigma_\gamma)$ and biases $\mathrm{b}(\boldsymbol{z}) \sim N(0, \sigma_b)$ are regularized to follow normal distributions, with mean zero, and learned variances $\sigma_\gamma$ and $\sigma_b$, respectively.

To further enforce the learning of batch-specific features in scMEDAL-RE, we add a RE batch classifier (Fig. 1 orange area, right side) which predicts cluster membership $\hat{\boldsymbol{z}}_L$ from the latent representation. By minimizing the prediction error the classifier, scMEDAL-RE is encouraged to produce latent representations and hence reconstructions that characterize the batch effects. Combining these components, the total loss function for the random effects subnetwork $\mathcal{L}_{RE}$ becomes:

$$\mathcal{L}_{RE}(\boldsymbol{x}, \widehat{\boldsymbol{x}}_R, \boldsymbol{z}, \hat{\boldsymbol{z}}_L,) = \lambda_{recon,R}\mathcal{L}_{MSE}(\boldsymbol{x}, \widehat{\boldsymbol{x}}_R) + \ \lambda_L \mathcal{L}_{CCE}(\boldsymbol{z}, \hat{\boldsymbol{z}}_L) + \lambda_K D_{KL}(q(U) \parallel p(U)) \quad (8)$$

where $\lambda_{MSE}$, $\lambda_{CCE_z}$and $\lambda_K$ are hyperparameters weighting each loss term. The first term encourages accurate reconstruction of the data, the second term ensures that the latent space captures batch-specific information, and the third term regularizes the model by aligning the learned distribution $q(U)$ with the prior $p(U)$, thereby minimizing overfitting.

Operationally, scMEDAL-RE conditions on the observed expression and batch labels, $(x, Z)$, and learns a batch-explicit latent representation $r = e_R(x; U(Z))$ the decoder then reconstructs $\widehat{\boldsymbol{x}}_R = h_R(r, Z;\ U)$. Because $Z$ is already provided directly to the decoder, copying $Z$ into $r$ does not improve the likelihood term and is discouraged by the KL regularizer in the ELBO. Gradient-based optimization therefore drives $r$ to encode within-batch, cell-specific residual structure—gene-level deviations conditional on $Z$—rather than duplicate one-hot encoded batch information. Consistent with this design, batch counterfactuals (holding $r$ fixed and swapping $Z$ at the decoder) yield coherent, batch-consistent shifts in reconstructed expression, indicating that $r$ captures batch-conditioned deviations, while the cross-entropy term $L_{CCE}(Z, \hat{Z}_L)$ encourages the latent to remain informative about batch without collapsing to an explicit copy of the label.

### 4.2.3 Overall framework

The overall framework, scMEDAL, integrates the fixed effects subnetwork (scMEDAL-FE) and the random effects subnetwork (scMEDAL-RE), which are optimized separately. As shown in Section 2.5, after inferring the embeddings $e_F(\boldsymbol{x}; \beta)$ and $e_R(\boldsymbol{x}; U(\mathbf{z}))$, we concatenate them and train a separate downstream classifier trained to infer a specific target label, $\hat{y}_M$ (e.g., the tissue, diagnosis, patient group or cell type). For completeness, a joint objective is provided in Eq. (9) which shows the combination of the FE and RE loss terms:

$$\lambda_{recon,F}\mathcal{L}_{MSE}(\boldsymbol{x}, \hat{\boldsymbol{x}}_F) + \lambda_{recon,R}\mathcal{L}_{MSE}(\boldsymbol{x}, \hat{\boldsymbol{x}}_R) + D_{KL}\big(q(U) \parallel p(\boldsymbol{U})\big) \; + \; \lambda_L\mathcal{L}_{CCE}(\boldsymbol{z}, \hat{\boldsymbol{z}}_L) \; - \; \lambda_A\mathcal{L}_{CCE}(\boldsymbol{z}, \hat{\boldsymbol{z}}_A) \quad (9)$$

In practice we typically find it more convenient to train each subnetwork separately --with its own objective: FE follows Eq. (2) (reconstruction with an adversarial, negatively weighted batch cross-entropy to yield a batch-agnostic latent), and RE follows Eq. (8) (reconstruction plus KL regularization with a positively weighted batch-prediction term to capture batch-dependent structure). This decoupled scheme improves scalability and stability on high-dimensional scRNA-seq data and enables plug-and-play integration of scMEDAL-RE with external batch-correction embeddings (e.g., scVI, Harmony, Scanorama, SAUCIE, scANVI) in the same downstream classification pipeline.

The scMEDAL hyperparameters are listed in Supplementary **Tables** S2–S4. All AE-based models were trained with the Adam optimizer with a learning rate of 0.0001 for 500 epochs. Early stopping, based on the reconstruction mean squared error (MSE) on validation data, was applied with a patience of 30 epochs as a regularization strategy to prevent overfitting.

## 4.3 Experiments and model comparisons

We rigorously evaluate the performance of the scMEDAL framework's components, including the primary innovation, scMEDAL-RE, the default batch correction approach, scMEDAL-FE, and the visualization tools within scMEDAL (e.g., for consistent genomap visualization). In particular, we benchmark scMEDAL-RE against widely used dimensionality-reduction and batch-correction approaches across diverse settings (autism, leukemia, cardiovascular), cell types, as well as technical and biological effects, providing a comprehensive assessment of the proposed framework. The first model we compare against is PCA[7], a standard and widely-used, linear method for dimensionality reduction in scRNA-seq analysis, whose properties are well understood, and it therefore served as a solid baseline for comparison. We then compare against a conventional autoencoder (AE)[40], which is a nonlinear dimensionality reduction technique widely applied in transcriptomics. The plain AE complements PCA method and serves as an ablation test for our default batch correction method, the autoencoder-based scMEDAL-FE model. We also extensively evaluate the complementarity of scMEDAL-RE to batch-correction methods that do not require cell type labels for training: scVI[14], Harmony[8], Scanorama[9], SAUCIE[27] as well as methods that exploit such labels when they are available. The following sections detail these experiments.

### 4.3.1 Experiment 1: Characterize cell type and batch separability and visualize the learned batch effects

To evaluate the performance and interpretability of the scMEDAL framework, quantitative and qualitative assessments of the learned representations of the two subnetworks are conducted. In the *quantitative assessment*, we measure cell-type and batch separability in each model's latent representation (for autoencoders, this is the bottleneck). scMEDAL-RE learns batch-specific latent spaces, whereas our default scMEDAL-FE, scVI[14], Harmony[8], Scanorama[9], SAUCIE[27] learn batch-invariant embeddings. We quantify this using three measures: ASW[28],CH[31] and the reciprocal of DB[33]. Higher clustering scores indicate greater (cleaner) clustering. For example, when clustered by cell type, higher clustering scores indicate better cell type separability.

To evaluate the batch-specific latent space of scMEDAL-RE, we compute its clustering by batch: *higher* scores relative to batch-suppression methods (and to AE/PCA baselines) indicate successful modeling of batch effects. Conversely, to assess batch suppression, we evaluate clustering by batch in each method's latent space, where *lower* scores denote more effective suppression. When batch–cell-type confounding is modest, stronger batch suppression typically coincides with higher clustering by cell type.

In the *qualitative* assessment, we characterize the batch-invariant and batch-specific latent spaces through 2D UMAP visualizations of the cells projected into each space. We color cells by cell type, batch, and batch-confounded biological factors (e.g., tissue, donor, diagnosis, patient group) to reveal how batch and biology relate to one another—patterns that batch-correction methods might otherwise obscure. We also characterize the latent spaces by generating 2D images, called genomaps, which depict gene expression for individual cells, with pixel intensity corresponding to gene expression level. Images are reconstructed from the batch-invariant latent space of scMEDAL-FE, and additional genomap images are constructed by leveraging the generative properties of the autoencoder, from the batch-specific latent spaces of scMEDAL-RE. Since this subnetwork learns to model the batch effects for each batch, we can use it to address retrospective "What if?" questions regarding what a cell's gene expression profile would look like if it had originated from a different batch, i.e., a different technical batch or biological donor. By intentionally setting the one-hot encoded batch vectors in the RE subnetwork, we project cells into different batches conditions. This projection capability enables exploration of the variability between different donors, or diagnoses, tissues, or technical data acquisition conditions. The results of this assessment are described in Section 2.5.

### 4.3.2 Experiment 2: Determine if scMEDAL-RE is complementary to supervised models that enhance cell type preservation

Many transcriptomics datasets include cell type labels from an external source. When such labels are available, this information can be leveraged by an *autoencoder classifier* (AEC) to further ensure preservation of cell type information. We construct the scMEDAL-FEC subnetwork, which predicts the one-hot encoded cell type labels $y$ by incorporating a categorical cross-entropy term $\mathcal{L}_{CCE}(y,\hat{y})$ into the fixed effects loss, weighted by $\lambda_y$. The updated fixed effects loss function $\mathcal{L}_{FEC}$ becomes:

$$\mathcal{L}_{FEC}(\boldsymbol{X},\boldsymbol{Z},y) = \lambda_{recon,F}\mathcal{L}_{MSE}\big(\boldsymbol{X},\widehat{\boldsymbol{X}}\big) - \lambda_A\mathcal{L}_{CCE}(\boldsymbol{z},\hat{\boldsymbol{z}}) + \lambda_y\mathcal{L}_{CCE}(y,\hat{y}) \quad (10)$$

We hypothesize that although an autoencoder-based cell-type classifier can improve cell-type embeddings, it will **not** disentangle batch–biology confounding (e.g., which is present across tissue, donor, and patient groups) and may therefore remove biologically meaningful signal that is entangled with batch. Moreover, we expect scMEDAL-RE to remain complementary in this setting. To test this central hypothesis, we replace the AE from Experiment 1 with an AEC (inspired by AutoClass[24]) ablation for scMEDAL-FEC, and add scANVI as a supervised comparator to scVI.

### 4.3.3 Experiment 3: Evaluate how the complementary nature of scMEDAL-RE embeddings enhances prediction performance

The aim of this experiment is to demonstrate that scMEDAL-RE's batch-specific latent spaces are complementary and can be combined with batch-agnostic embeddings from various correction methods to improve cell classification accuracy. This experiment explores the extent to which quantitatively modeling batch effects, rather than discarding them as other batch correction methods do, can improve predictions about the cells. To assess the complementary nature of scMEDAL-RE, we use the Experiment 1 and 2 embeddings from PCA, scMEDAL-FE, scVI[14], Harmony[8], Scanorama[9] and SAUCIE[27], scMEDAL-FEC, scANVI[15], and compare the predictive power of each method's latent space alone versus the concatenation of that embedding of the cell, with the scMEDAL-RE's cell embedding.

The Random Forest model was chosen because it is interpretable, effectively handles both linear and nonlinear relationships, and it is not prone to overfitting. We used the same 5-fold cross-validation partitioning scheme described for preprocessing the data input for the scMEDAL subnetworks. The PCA latent space was also utilized as a comparative baseline. This approach enabled us to assess how the integration of scMEDAL-RE enhances classification performance compared to other methods alone. In the Healthy Heart dataset we classified tissue, in the AML dataset, we also conducted experiments using patient group as a target, while in the ASD dataset we used disease diagnosis: autism spectrum disorder or typically developing (control) as an additional target. Since all three datasets[41,42,44] included cell type labels, we used cell type as the secondary target for classification for PCA, scMEDAL-FE, scVI[14], Harmony[8], Scanorama[9] and SAUCIE[27].

## 4.4 Performance Metrics

The core metric used to guide our scMEDAL training is the validation total loss for each of the subnetworks which is described in equations 2, 8, and 10. Additionally, clustering metrics were used to evaluate the biological separability, batch correction, and batch modeling.

### 4.4.1 Clustering performance metrics

To evaluate cell type separability, batch correction, and batch modeling capabilities[29,30,32], we utilize three clustering metrics: the Average Silhouette Width (ASW)[28], Calinski-Harabasz (CH) index[31], and Davies-Bouldin (DB) index[33]. The ASW is a widely used metric[19,23,34-36] that quantifies how well each cell is assigned to its cluster compared to other clusters. The DB index quantifies how much clusters overlap (lower values indicate less overlap), while the CH index measures between-cluster *versus* within-cluster variation (higher values suggest better-defined clusters). Unlike the Calinski-Harabasz (CH) index and the Davies-Bouldin (DB) index—which assume convex or spherical clusters and focus on cluster variance structure and overlap, respectively—ASW does not rely on such assumptions.[37] Therefore in this work we choose ASW to serve as our primary metric because it is more robust to clusters of varying shapes compared to CH and DB[37] and its values range from -1 to 1, facilitating comparisons across methods and datasets. We still including the CH and DB indices, thereby ensuring a comprehensive evaluation from different perspectives: CH captures variance structure and compactness, while DB highlights the degree of overlap between clusters and to facilitate comparison to other studies. See Supplementary information section 2.6 for the mathematical formulation of each metric.

### 4.4.2 Classification performance metrics

To evaluate the classification performance of predicting labels from the scMEDAL latent spaces, we calculated accuracy, balanced accuracy, and chance accuracy. For chance accuracy, we employed scikit-learn's *Dummy Classifier* with the *stratified* strategy, which generates random predictions while maintaining the original class distribution in the dataset.[7] This provides a more realistic baseline for comparison, as it accounts for the imbalanced nature of the classes rather than assuming uniform distribution, thus offering a more appropriate measure of chance-level performance. All these metrics range between 0 and 100%. Balanced accuracy accounts for class imbalance, which is crucial in our datasets due to multiple cell type classes with varying abundances. In the results section, we report results including ASW, classification accuracy, balanced accuracy and chance accuracy on the test data, which is not used for model training nor model selection.

## 4.5 Hyperparameter optimization

There are three types of hyperparameters that shape the proposed framework, including: (1) dimensions of the autoencoder neural networks (e.g., the fixed and random effects subnetworks), (2) choices that govern the optimization of the models' loss function, and (3) the weights that balance the individual terms of the overall loss function. The proposed framework does not require precise tuning of all these design decisions. In fact, reasonable estimates were made for the first two types of hyperparameters, and this sufficed to obtain good performance and convergence properties for all five of the autoencoder models (AE, AEC, scMEDAL-FE, scMEDAL-FEC, scMEDAL-RE) across all three datasets. This underscores the appropriateness of the chosen approach for scRNA-seq analysis. All the autoencoder models, for all datasets, employed three hidden layers with 512 neurons in the first layer, 132 in the second layer, and two neurons in the bottleneck (third) layer. Weight tying between the encoder and decoder components was used for all autoencoders except for scMEDAL-RE, where it was not needed. For the loss function optimization, all models used the Adam optimizer with a fixed learning rate of $lr = 0.0001$. Additionally, for all datasets and models, early stopping based on validation loss was utilized as a convenient and reliable approach for regularization and the prevention of overfitting, ensuring robust and generalizable models. Taken together, this suggests the broad generalizability of the proposed approach and attests to the fairness of the comparison between the proposed approach and the AE and AEC models, as all models employed the same network depth.

In practice we find that only the scaling weights of the terms in the loss function, needs to be optimized per dataset (see Supplementary **Tables** S2–S4 for the values employed) and this optimization is straightforward using the approach we will now describe. We begin by partitioning the data into training, validation and test sets using a 5-fold, multiple hold out, cross-validation strategy. Proper optimization is then achieved by ensuring that the loss terms, such as the reconstruction loss, have roughly the same order of magnitude as the other loss terms during the initial training epochs. Whether the terms are balanced can be readily inspected by constructing training curves with separate curves plotted for the contribution of each loss term, as well as a curve for the total loss function (See Supplementary **Figs**. S5-S10). Weights can then be readily adjusted so that each of the terms (adversarial, cell type classifier, batch classifier, and KL losses) is roughly the same magnitude as the reconstruction loss during the initial epochs of training. Maintaining this balance facilitates proper optimization across all components throughout training. Note that when constructing the AEC and scMEDAL-FEC models, additional information is available in the form of cell type labels. In that scenario it is prudent to additionally monitor cell type separability in the latent spaces as measured by cell type ASW to further optimize weight selection. For model selection, the hyperparameter weights that attained the highest performance on the validation folds are chosen. To minimize inflation bias, the final model's performance is evaluated on the held-out test folds that were not used in training or validation. In general, using the strategy described above, the weights for the loss terms for all models are readily tuned to balance the terms, leading to the rapid convergence of the model weights through total loss function optimization.

## 4.6 Visualization methods

To provide a comprehensive evaluation of our model's performance, two visualization techniques were employed: Uniform Manifold Approximation and Projection (UMAP)[38] and genomaps[45]. The UMAP approach is used to visualize learned latent spaces for an intuitive understanding of the spaces and as a complement to the quantitative evaluation metrics: DB index, CH index, and ASW. This combined approach allows us to better assess both cell-type signal enhancement and batch separability. The genomap approach was used to construct intuitive 2D image representations of each cell's gene expression profile. This approach arranges gene expression levels into a 2D polar coordinate grid according to their gene-to-gene interactions. In this coordinate space, the origin of the image is at the center of the image. Genes are placed with a distance $\rho$ from the image origin (center of image) proportional to their level of interaction with other genes. Thus, genes positioned at the same distance, $\rho$, from the center have equivalent levels of gene-to-gene interaction. The angular component, $\theta$, is arbitrary. The intensity of the pixel in the image representation corresponds to the gene expression level. Hence, if there are many highly expressed genes with the same gene-to-gene interaction, a bright annulus will appear in the image, whose radius corresponds to the common level of interaction. Islam and Xing et al. (2023) demonstrated that genomaps can create gene expression patterns specific to cell types[45]. Importantly, genomap representations preserves the original gene expression values while organizing the genes spatially, allowing for a clear visualization of the relationships between genes and providing insights into batch effects when combined with our generative modeling framework. We use genomap visualization to gain greater understanding of cell-to-cell differences (e.g., malignant vs healthy cells) in the original data and in the batch agnostic representation learned by the fixed effects portion (scMEDAL-FE) of our framework and batch-to-batch variability learned by the random effects portion (scMEDAL-RE) of our framework.

## 5. Code and data availability

We will make all code publicly available upon journal publication. Gene expression matrices and cell annotations for the AML dataset were obtained from the Gene Expression Omnibus (GEO) under accession number GSE116256[44]. The Healthy Heart dataset was acquired from Yu et al. (2023) through figshare (DOI: https://doi.org/10.6084/m9.figshare.20499630.v2)[41], and it may also be obtained from the Heart Cell Atlas (https://www.heartcellatlas.org/)[39]. The ASD dataset was obtained from the UCSC Cell Browser (https://autism.cells.ucsc.edu )[42,65]

## 6. Acknowledgements

This study was supported by the NIH R01 Grant "Correcting Biases in Deep Learning Models" (5R01GM144486-02). We also acknowledge the BioHPC supercomputing facility at the Lyda Hill Department of Bioinformatics, UT Southwestern Medical Center, TX, for providing computational resources (URL: https://portal.biohpc.swmed.edu). Special thanks to Dr. Prapti Mody for her invaluable feedback on this work.

## 7. Ethics Approval

This study did not involve the collection of data from human participants, animal subjects, or biological samples. All datasets used in this work were previously published as part of their respective studies. As the study is purely computational, no ethical approval was required.

## 8. Competing Interests

The authors declare no competing interests.

# scMEDAL for the interpretable analysis of single-cell transcriptomics data with batch effect visualization using a deep mixed effects autoencoder

Aixa X. Andrade[1], Son Nguyen[1], Austin Marckx and Albert Montillo[1]*
[1]Lyda Hill Department of Bioinformatics
University of Texas Southwestern Medical Center
Dallas, TX 75390, USA
*Corresponding author: Albert Montillo[1], Ph.D.; Email: albert.montillo@utsouthwestern.edu

# Supplementary information

# 1. Supplementary materials

**Table** S1 shows the details of the datasets used to evaluate the scMEDAL subnetworks as well as the baseline models (PCA and the AE and AEC) and comparable batch correction models. Data preprocessing steps are described in supplemental section 2.2

**Table S1**. Datasets used to evaluate scMEDAL models.

| | *Healthy Heart*[1,2] | *ASD*[3] | *AML*[4] |
|---|---|---|---|
| Description | Heart cells from multiple tissues | ASD and TD (controls) | AML and healthy subjects |
| Data type | Single cell | Single nucleus | Single cell |
| Total cells (pre-filtering) | 486,134 | 104, 559 | 41,090 |
| Total cells (post-filtering) | 486,134 | 104, 559 | 38,417 |
| Total genes | 33,538 | 36,501 | 27,899 |
| Highly variable genes | 3,000 | 2,916 | 2,916 |
| Number of cell types | 12 + Not Assigned cell type | 17 | 6 malignant + 15 healthy |
| Batch effect type | Technical batch (z =147) | Biological donor (z = 31) | Technical batch (z=19) |

# 2. Supplementary methods

## 2.1. Supplementary figures of the scMEDAL architecture

### Random effects *Bayesian* layer block

**Fig.** S1 depicts the architecture of the Random Effects Dense (REDEN) Bayesian layer block, which is used to construct the Bayesian layers of the scMEDAL-RE subnetwork.

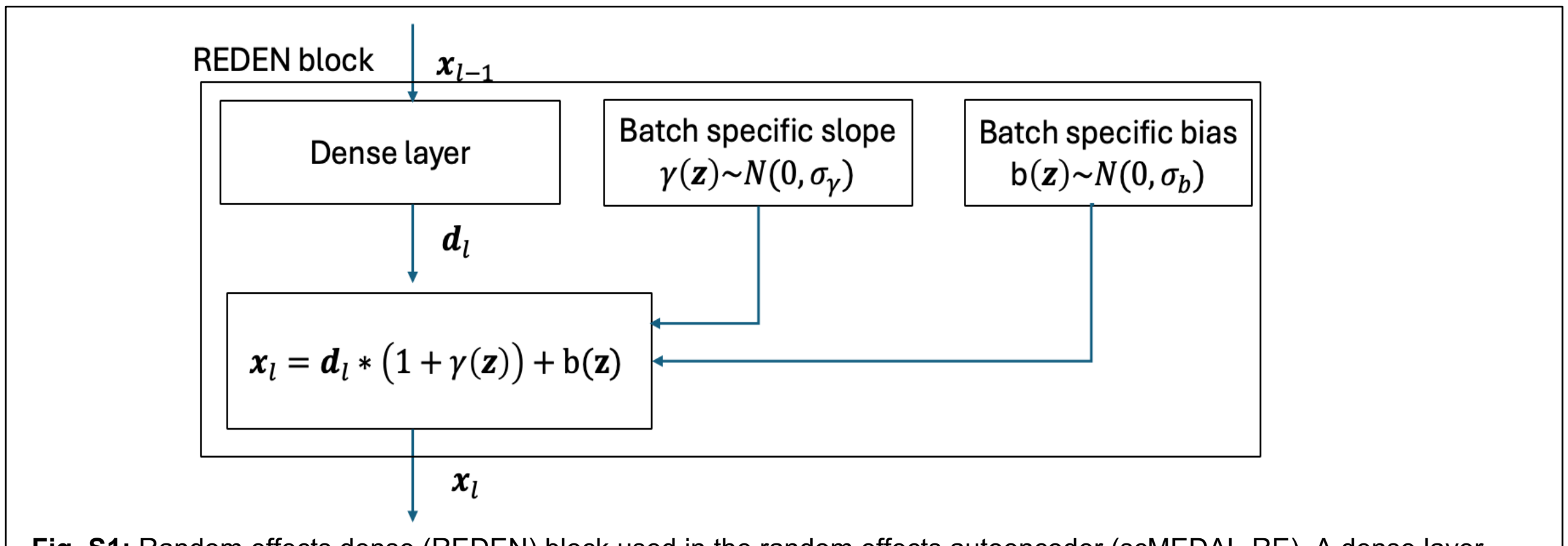

**Fig. S1:** Random effects dense (REDEN) block used in the random effects autoencoder (scMEDAL-RE). A dense layer with $n_l$ neurons transforms an input representation from the preceding layer, $x_{l-1}$, into an intermediate output tensor, $d_l$. A batch-specific scale *γ(**z**)* and bias *b(**z**)* are then applied to each of the *f* feature maps. These batch-specific slopes and biases are regularized to follow normal distributions, with mean zero, and learned variances $\sigma_\gamma$ and $\sigma_b$, respectively.

## 2.2. Data preprocessing

We used standard preprocessing steps from Yu et al. , 2023[5] which included: 1) filtering out cells exhibiting expression in fewer than ten genes and removal of genes detected in fewer than three cells, 2) normalizing total Unique Molecular Identifiers (UMI) counts per cell to 10,000, 3) log-transforming the data to stabilize variance $\log(X+1)$, and 4) selecting the top highly variable genes (HVGs).

Since the ASD and Healthy Heart datasets had previously undergone quality control, no additional cells were removed during the filtering step. For the AML dataset, additional quality control measures were implemented. Cells with undefined or missing cell type annotations were excluded. Furthermore, samples identified by Dai et al., (2021) as having ambiguous annotations—specifically AML314, AML371, AML722B, and AML997[6]—were removed from the original 16 AML donors to ensure the dataset's integrity, resulting in 12 AML donors available for analysis. Including 5 healthy donors and 2 cell lines, the dataset comprised a total of 19 batches. The ASD dataset had previously undergone a log2 transformation as part of the original preprocessing by van Galen et al. (2019). To maintain consistency across all datasets, this transformation was reversed by applying an exponential function (base 2) to the data after adding 1 to each value, thereby returning the data to its original scale. Following this reversal, the standard preprocessing pipeline was applied.

The datasets were then partitioned through 5-fold cross-validation, stratified by batch and cell type in all three datasets. For each split, three folds were used for training, one for validation, and one for testing. The data were loaded for each subset and scaled using min-max scaling before model fitting. To reduce compute time for the ASW calculations, we used a random sample of 10,000 cells when datasets exceeded this size in training, validation, or testing subsets.

## 2.3. Hyperparameters and model training details

### 2.3.1. Hyperparameters for AE, AEC, scMEDAL-FE, scMEDAL-FEC, and scMEDAL-RE models

Section 4.5 of the main paper describes the hyperparameter optimization process describes the hyperparameter-optimization procedure for scMEDAL-FE (Equation 2), scMEDAL-RE (Equation 8) and scMEDAL-FEC (Equation 10). In the following **Tables** S2-S4, we describe the values of the hyperparameter that are found to optimize the **AE, AEC, scMEDAL-FE, scMEDAL-FEC**, and **scMEDAL-RE** models in three datasets: Healthy Heart, the Autism Spectrum Disorder (ASD), and Acute Myeloid Leukemia (AML).

**Table S2.** Hyperparameters used for training PCA, AE and scMEDAL subnetworks in the **Healthy Heart dataset**.

| Hyperparameter | AE | AEC | scMEDAL-FE | scMEDAL-FEC | scMEDAL-RE |
|---|---|---|---|---|---|
| $\lambda_{recon,F}$ | 1 | 100 | 600 | 2000 | - |
| $\lambda_{recon,R}$ | - | - | - | - | 110 |
| $\lambda_y$ | - | 0.1 | - | 1 | - |
| $\lambda_A$ | - | - | 1 | 1 | - |
| $\lambda_L$ | - | | | | 0.1 |
| $\lambda_K$ | - | | | | 1.00E-05 |

**Table S3.** Hyperparameters used for training PCA, AE and scMEDAL subnetworks in the **ASD dataset**.

| Hyperparameter | AE | AEC | scMEDAL-FE | scMEDAL-FEC | scMEDAL-RE |
|---|---|---|---|---|---|
| $\lambda_{recon,F}$ | 1 | 100 | 2000 | 2000 | - |
| $\lambda_{recon,R}$ | - | - | - | - | 110 |
| $\lambda_y$ | - | 0.1 | - | 1 | - |
| $\lambda_A$ | - | | 1 | 1 | - |
| $\lambda_L$ | - | | | | 0.4 |
| $\lambda_K$ | - | | | | 1.00E-05 |

**Table S4.** Hyperparameters used for training PCA, AE and scMEDAL subnetworks **AML dataset**.

| Hyperparameter | AE | AEC | scMEDAL-FE | scMEDAL-FEC | scMEDAL-RE |
|---|---|---|---|---|---|
| $\lambda_{recon,F}$ | 1 | 100 | 1000 | 2000 | - |
| $\lambda_{recon,R}$ | - | - | - | - | 110 |
| $\lambda_y$ | - | 0.6 | - | 3 | - |
| $\lambda_A$ | - | - | 1 | 0.5 | - |
| $\lambda_L$ | - | | | | 0.1 |
| $\lambda_K$ | - | | | | 1.00E-05 |

### 2.3.2. Hyperparameter Optimization (HPO) for AEC and scMEDAL-FEC models

Hyperparameter optimization (HPO) was conducted per dataset to tune the parameters using the held-out validation data (not used for model training). This tuning adjusts the weights of the reconstruction, adversarial, classification (cell type for scMEDAL-FE and batch for scMEDAL-RE), and Kullback–Leibler divergence loss function terms so that they are on a similar scale, allowing all terms to guide the model fitting, and optimizing performance on the validation data. However, when cell type information is available from an external source (as described in Section 2.6 of the main paper) this can be used to further guide hyperparameter tuning. In this scenario, the ASW for cell type can be used to further refine the loss function weights. Both batch and cell-type ASW on the AEC and scMEDAL-FEC models can be evaluated during the training for 500 epochs with early stopping, prioritizing the cell-type ASW.

### 2.3.3. Hyperparameters for comparable models

For scVI[7], we used a 50-dimensional latent space (for comparison with the same-sized scMEDAL-FE and RE latent spaces) with two encoder/decoder layers of 132 hidden units each, a zero-inflated negative binomial (ZINB) gene-expression likelihood, and gene-level dispersion. scANVI[8] employed the same architectural and likelihood settings as scVI. For SAUCIE[9] (autoencoder-based), we used a three-layer encoder (512→132→50), taking the 50-unit layer as the latent space; clustering and intracluster-distance penalties were disabled ($\lambda_c = 0.0$, $\lambda_d = 0.0$), and models were trained for 50 epochs with a learning rate of $1\times10^{-4}$ to avoid overfitting. Batch regularization was deliberately mild, with $\lambda_b$ = 0.001 for ASD and AML, and a slightly lower $\lambda_b$ = 0.0005 for Healthy Heart to prevent over-penalization under its more complex, multi-factor batch structure.

### 2.3.4. Training notes for comparable models

For scVI[7], each fold was trained on the training partition using batch as the covariate. We then encoded the training, validation, and test sets separately to obtain latent representations and normalized expression—no cross-split merge was needed. scVI uses an internal 10% validation split with early stopping, so the data used during optimization differed from the predefined fold validation set; we report results on the original scMEDAL train/validation/test partitions.

For scANVI[8] , we followed the same per-fold protocol (no merge across splits) but trained in a supervised manner with annotated cell-type labels guiding the latent space. We initialized from an scVI model fit on the training partition (batch key = batch; label = cell type), then converted and fine-tuned it as scANVI (batch size 512; up to 500 epochs; early stopping with patience = 30; internal 10% validation split). After training, we computed latent representations and normalized expression for the original scMEDAL partitions.

SAUCIE[9] was trained with the same 5-fold splits used for scMEDAL. By contrast, Scanorama[10] and Harmony[11] require a merge–integrate–split workflow because they perform joint integration rather than a per-split fit–transform. For each fold, we merged the train/validation/test partitions, computed 50 principal components (PC), performed the integration in PCA space, and then restored the original partitions while preserving cell order. For Scanorama, we additionally sorted cells by batch to make batches contiguous before integration. For Harmony, contiguous ordering was unnecessary; Harmony iteratively adjusts the PC embedding with batch-specific correction terms.

## 2.4. UMAP visualization parameters

**Table** S5 describes the minimum distance and nearest neighbor's choices for UMAP visualizations presented in the main paper in sections 2.2, 2.3, 2.4, and 2.6.

**Table S5.** UMAP Visualization parameters.

| Dataset | UMAP parameters | All models |
|---|---|---|
| Healthy Heart | Min distance | 0.5 |
| | Nearest Neighbors | 15 |
| ASD | Min distance | 0.5 |
| | Nearest Neighbors | 15 |
| AML | Min distance | 0.5 |
| | Nearest Neighbors | 15 |

## 2.5. Hardware and software used for model implementation and training

All deep learning models were developed and trained on Nvidia Tesla V100 GPUs with 32 GB of memory and Tesla P4 GPUs with 8 GB of memory. The software stack for model training included Python 3.8.20, TensorFlow 2.13.1[12], TensorFlow Probability 0.21.0[12-14], scikit-learn 1.3.2, and Scanpy 1.9.8. Performance measures, including ASW, DB, CH, accuracy, balanced accuracy, and chance accuracy, were computed using the Scikit-learn library[15]. The PCA model was also implemented through the Scikit-learn library. scRNA-seq preprocessing and UMAP computations were performed with Scanpy[16]. Genomaps were generated using genomap 1.3.6 and the Mann-Whitney U test was implemented using the using the *stats* module from SciPy 1.10.1[17] and the linear mixed-effects models were implemented in the Statsmodels 0.14.1 library[18].

The comparable models were implemented as follows: scVI and scANVI used Python 3.12.10 with scvi-tools 1.3.0 and PyTorch 2.2.0+cu121; Harmony and Scanorama were accessed via the Scanpy external toolkit; and SAUCIE was run with Scanpy 1.7.2, scikit-learn 0.24.2, and tensorflow-gpu 1.12.0.

For more details on the software used for model training, plot creation, and generation of visualizations, please refer to the code repository described in the *Code and data availability* section 5.

## 2.6. Clustering metric definitions

We use the following metrics to quantify the clustering (separability) of the cells: the Average Silhouette Width (ASW)[19], Calinski-Harabasz (CH) index[20], and Davies-Bouldin (DB) index[21]. The *Silhouette Width* quantifies how well each cell fits within its assigned cluster compared to the other clusters. It is computed using two distances: the within-cluster distance ($a_i$) and the between-cluster distance ($b_i$). The within-cluster distance, $a_i$, represents the average distance between a cell $i$ and all other cells in its cluster. This distance is defined as:

$$a_i = \frac{1}{|C_i| - 1} \sum_{\{j \in C_i, j \neq i\}} d(i,j)$$

where $C_i$ is the set of cells in the cluster of cell $i$, $|C_i|$ is the total number of cells in this cluster, and $d(i,j)$ is the distance between cell $i$ and cell $j$ within the cluster. In contrast, the between-cluster distance, $b_i$, is the minimum of the average distance between cell $i$ and cells from other clusters. It is defined as:

$$b_i = min_{\{k \neq i\}} \frac{1}{|C_k|} \sum_{\{j \in C_k\}} d(i,j)$$

where $C_k$ is the set of cells assigned to cluster $k$ which is different from the cluster of cell $i$, while $|C_k|$ is the total number of cells in cluster $k$. The silhouette coefficient for each cell $i$ is calculated as:

$$s_i = \frac{b_i - a_i}{max(b_i, a_i)}$$

where a higher $s_i$ value (closer to 1.0) indicates that the cell is well-clustered, being more similar to cells within its own cluster than to those in any other cluster. The *ASW* is defined as the average of $s_i$ over all cells and provides an overall measure of clustering quality.

The *Davies–Bouldin* (DB) *index* quantifies the extent to which clusters overlap. For each cluster, we compare its average within-cluster dispersion to the distance between its centroid and the centroid of the most similar (closest) cluster. Consequently, a lower DB index value indicates less overlap and better separation between clusters. The DB index is computed as:

$$DB = \left(\frac{1}{K}\right) \sum_{\{k=1\}}^{K} D_k$$

where $K$ is the number of clusters, and $D_k$ for cluster $k$ is the ratio, $R_{kl}$ for the most similar cluster $l$ to $k$, $(l \neq k)$. $D_k$ is expressed as:

$$D_k = max_{\{l \neq k\}} R_{kl}$$

where the ratio $R_{kl}$ measures how similar are the clusters $k$ and $l$. $R_{kl}$ is computed as:

$$R_{kl} = \frac{(q_k + q_l)}{M_{kl}}$$

where $q_k$ is the is the average distance from each cell in cluster $k$ to the centroid of cluster $k$, $q_l$ is the corresponding average distance for cluster $l$, and $M_{kl}$ is the distance between the centroids of clusters $k$ and $l$. A larger $M_{kl}$ indicates better separation between the two clusters.

The *Calinski-Harabasz* (CH) index measures the ratio of between-cluster variance to within-cluster variance, with higher values indicating more distinct and cohesive clusters. It is calculated as

$$CH = \frac{\sum_{\{k=1\}}^{K} n_k * || \mu_k - \mu ||^2}{\sum_{\{k=1\}}^{K} \sum_{\{i=1\}}^{n_k} || p_i - \mu_k ||^2} * \left[\frac{N-K}{K-1}\right]$$

where $n_k$ is the number of points in the $k^{th}$ cluster, $\mu_k$ is the centroid of the $k^{th}$ cluster, $\mu$ is the centroid of the entire data (all the cells), $p_i$ is the position (coordinates) of the cell $i$, $N$ is the total number of cells and $K$ is the number of clusters.

To align the interpretation of the DB index with the CH index and ASW, we used the reciprocal (1/DB), ensuring that higher scores consistently indicate better clustering quality. When measuring the clustering by cell type, higher scores signify improved separability in the latent space, enhancing the cell type signal—an objective of batch correction. Conversely, when measuring the clustering by batch label, lower scores indicate more effective batch correction, as it indicates that the batches become less distinguishable, which is another objective of batch correction. Additionally, higher batch scores in the batch-specific latent space of the random effects subnetwork indicates effective modeling of the batch effects and capture of the between-batch variance in gene expression data.

# 3. Supplementary results

## 3.5. Confounding within the AML dataset

In **Fig**. S2 the representation of donors across cell types in AML is shown graphically. We observe that not all donor samples contain cells from every cell type, making this dataset a particularly challenging and interesting one for batch correction due to potential confounding between donor identity and cell type.

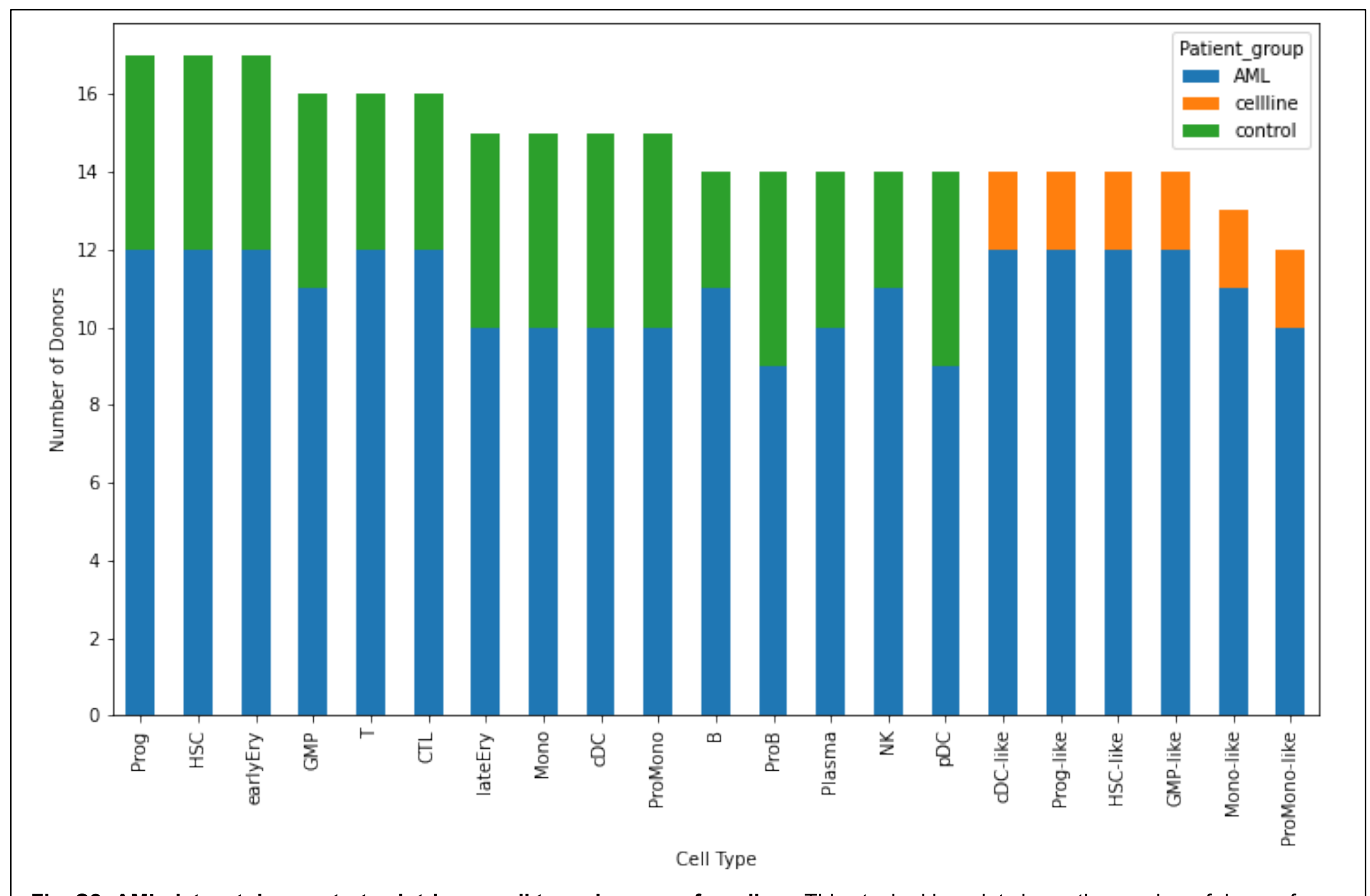

**Fig. S2: AML dataset demonstrates intrinsec cell type-donor confounding.** This stacked bar plot shows the number of donors for each cell type (y-axis), grouped by patient group, with cell types displayed along the x-axis. The colors indicate the donor patient group.

## 3.6. Supplementary results for PCA, AE, scMEDAL, scVI, Harmony, Scanorama and SAUCIE models

### 3.6.1. CH and 1/DB scores

In **Tables** S6-S8, scores are shown from the Calinski-Harabasz (CH) clustering index and the reciprocal of the Davies-Bouldin (DB) clustering index (1/DB), as these complement the primary metric, ASW, used in the main paper. These scores confirm the ASW results. In particular, the scMEDAL-RE subnetwork consistently captures batch specific information as designed, demonstrating significantly more batch information (higher scores) than any of the other models, as measured with both the CH and 1/DB indices, across all three datasets ($4^{th}$ row, first 2 columns of **Tables** S6-S8). Meanwhile, the scMEDAL-FE subnetwork, scVI[7], Harmony[11], Scanorama[10] and SAUCIE[9] suppresses batch specific information, demonstrating lower CH and 1/DB scores (less batch contamination) than the baseline models PCA and AE in all three datasets (**Tables** S6-S8). In terms of preserving cell type information, across all datasets, the batch correction methods outperform the baseline models in Healthy Heart (**Table** S6), mixed results in ASD (**Tables** S7) and deteriorating results in AML due to stronger batch cell-type confounding (**Table** S8, **Fig.** S2).

**Table S6.** 1/DB and CH scores (Mean and 95% CI across 5 folds) for batch and cell type separability in the latent spaces of the **Healthy Heart dataset**, using PCA, AE, scMEDAL-FE, scMEDAL-RE, scVI, Harmony, Scanorama and SAUCIE models. Bolded scores indicate the highest separability for each metric.

| | batch | | | | | | cell type | | | | | |
|---|---|---|---|---|---|---|---|---|---|---|---|---|
| | 1/DB | | | CH | | | 1/DB | | | CH | | |
| | mean | 95% CI | | mean | 95% CI | | mean | 95% CI | | mean | 95% CI | |
| PCA (Baseline) | 0.14 | 0.13 | 0.14 | 47.25 | 46.55 | 47.96 | 0.40 | 0.38 | 0.42 | 691.12 | 678.39 | 703.84 |
| AE | 0.13 | 0.13 | 0.14 | 50.23 | 33.88 | 66.59 | 0.38 | 0.34 | 0.41 | 796.46 | 698.29 | 894.64 |
| scMEDAL-FE | 0.10 | 0.10 | 0.11 | 28.55 | 14.96 | 42.14 | 0.37 | 0.35 | 0.39 | 762.33 | 612.55 | 912.10 |
| scMEDAL-RE | **0.42** | **0.37** | **0.47** | **333.73** | **239.78** | **427.68** | 0.20 | 0.18 | 0.21 | 268.98 | 207.41 | 330.54 |
| scVI | 0.06 | 0.05 | 0.06 | 5.23 | 4.56 | 5.90 | **0.55** | **0.45** | **0.64** | 2010.52 | 1848.34 | 2172.70 |
| Harmony | 0.11 | 0.06 | 0.15 | 31.92 | 11.70 | 52.13 | 0.27 | 0.13 | 0.41 | 875.22 | 639.57 | 1110.87 |
| Scanorama | 0.12 | 0.10 | 0.14 | 48.17 | 45.76 | 50.59 | 0.40 | 0.38 | 0.42 | 748.77 | 674.34 | 823.20 |
| SAUCIE | 0.06 | 0.05 | 0.07 | 15.74 | 6.10 | 25.37 | 0.20 | 0.13 | 0.27 | **2354.75** | **459.41** | **4250.10** |

**Table S7.** 1/DB and CH scores (Mean and 95% CI across 5 folds) for batch and cell type separability in the latent spaces of the **ASD**, using PCA, AE, scMEDAL-FE, scMEDAL-RE, scVI, Harmony, Scanorama and SAUCIE models. Bolded scores indicate the highest (best) separability for each metric.

| | batch | | | | | | cell type | | | | | |
|---|---|---|---|---|---|---|---|---|---|---|---|---|
| | 1/DB | | | CH | | | 1/DB | | | CH | | |
| | mean | 95% CI | | mean | 95% CI | | mean | 95% CI | | mean | 95% CI | |
| PCA (Baseline) | 0.09 | 0.09 | 0.10 | 25.74 | 25.16 | 26.33 | **0.46** | **0.45** | **0.47** | 1440.09 | 1427.94 | 1452.23 |
| AE | 0.09 | 0.09 | 0.10 | 24.67 | 24.05 | 25.29 | 0.44 | 0.43 | 0.44 | 1523.30 | 1432.99 | 1613.62 |
| scMEDAL-FE | 0.08 | 0.08 | 0.08 | 16.28 | 14.71 | 17.85 | 0.36 | 0.35 | 0.37 | 982.21 | 952.59 | 1011.84 |
| scMEDAL-RE | **0.30** | **0.27** | **0.33** | **144.72** | **118.01** | **171.42** | 0.30 | 0.28 | 0.31 | 436.08 | 376.06 | 496.11 |
| scVI | 0.06 | 0.05 | 0.06 | 11.83 | 10.97 | 12.69 | 0.43 | 0.39 | 0.46 | **4518.33** | **4303.85** | **4732.81** |
| Harmony | 0.07 | 0.04 | 0.10 | 18.38 | 8.73 | 28.04 | 0.45 | 0.43 | 0.46 | 1479.23 | 1393.49 | 1564.98 |
| Scanorama | 0.08 | 0.05 | 0.10 | 20.37 | 12.96 | 27.78 | 0.43 | 0.39 | 0.47 | 2136.42 | 1163.33 | 3109.51 |
| SAUCIE | 0.08 | 0.08 | 0.08 | 12.49 | 7.76 | 17.22 | 0.45 | 0.41 | 0.48 | 1716.57 | 1295.78 | 2137.35 |

**Table S8.** 1/DB and CH scores (Mean and 95% CI across 5 folds) for batch and cell type separability in the latent spaces of the **AML dataset**, using PCA, AE, scMEDAL-FE, scMEDAL-RE, scVI, Harmony, Scanorama and SAUCIE models. Bolded scores indicate the highest separability for each metric.

| | batch | | | | | | cell type | | | | | |
|---|---|---|---|---|---|---|---|---|---|---|---|---|
| | 1/DB | | | CH | | | 1/DB | | | CH | | |
| | mean | 95% CI | | mean | 95% CI | | mean | 95% CI | | mean | 95% CI | |
| PCA (Baseline) | 0.20 | 0.20 | 0.20 | 109.69 | 108.51 | 110.86 | 0.27 | 0.27 | 0.27 | 329.80 | 326.81 | 332.78 |
| AE | 0.19 | 0.19 | 0.20 | 117.36 | 112.13 | 122.59 | 0.27 | 0.26 | 0.27 | 400.15 | 355.55 | 444.75 |
| scMEDAL-FE | 0.17 | 0.17 | 0.18 | 93.79 | 90.59 | 96.99 | 0.25 | 0.25 | 0.26 | 323.70 | 305.35 | 342.05 |
| scMEDAL-RE | **1.50** | **1.29** | **1.71** | **7097.61** | **6162.38** | **8032.85** | 0.18 | 0.17 | 0.19 | 139.51 | 131.70 | 147.32 |
| scVI | 0.05 | 0.05 | 0.06 | 14.24 | 11.88 | 16.60 | 0.29 | 0.24 | 0.34 | **885.58** | **812.33** | **958.83** |
| Harmony | 0.14 | 0.05 | 0.22 | 71.37 | 22.16 | 120.58 | 0.21 | 0.13 | 0.29 | 298.48 | 263.40 | 333.57 |
| Scanorama | 0.16 | 0.10 | 0.21 | 85.09 | 53.43 | 116.76 | 0.25 | 0.22 | 0.28 | 393.25 | 253.51 | 532.99 |
| SAUCIE | 0.16 | 0.15 | 0.18 | 109.44 | 91.25 | 127.62 | **0.30** | **0.29** | **0.32** | 638.71 | 493.41 | 784.01 |

### 3.6.2. Healthy Heart latent spaces colored by protocol and donor ID.

**Figs.** S3 and S4 show the latent spaces through UMAP visualizations for the Healthy Heart dataset for when colored by protocol and donor ID. We observe that different methods show different degree of cell mixing between protocols and donor IDs.

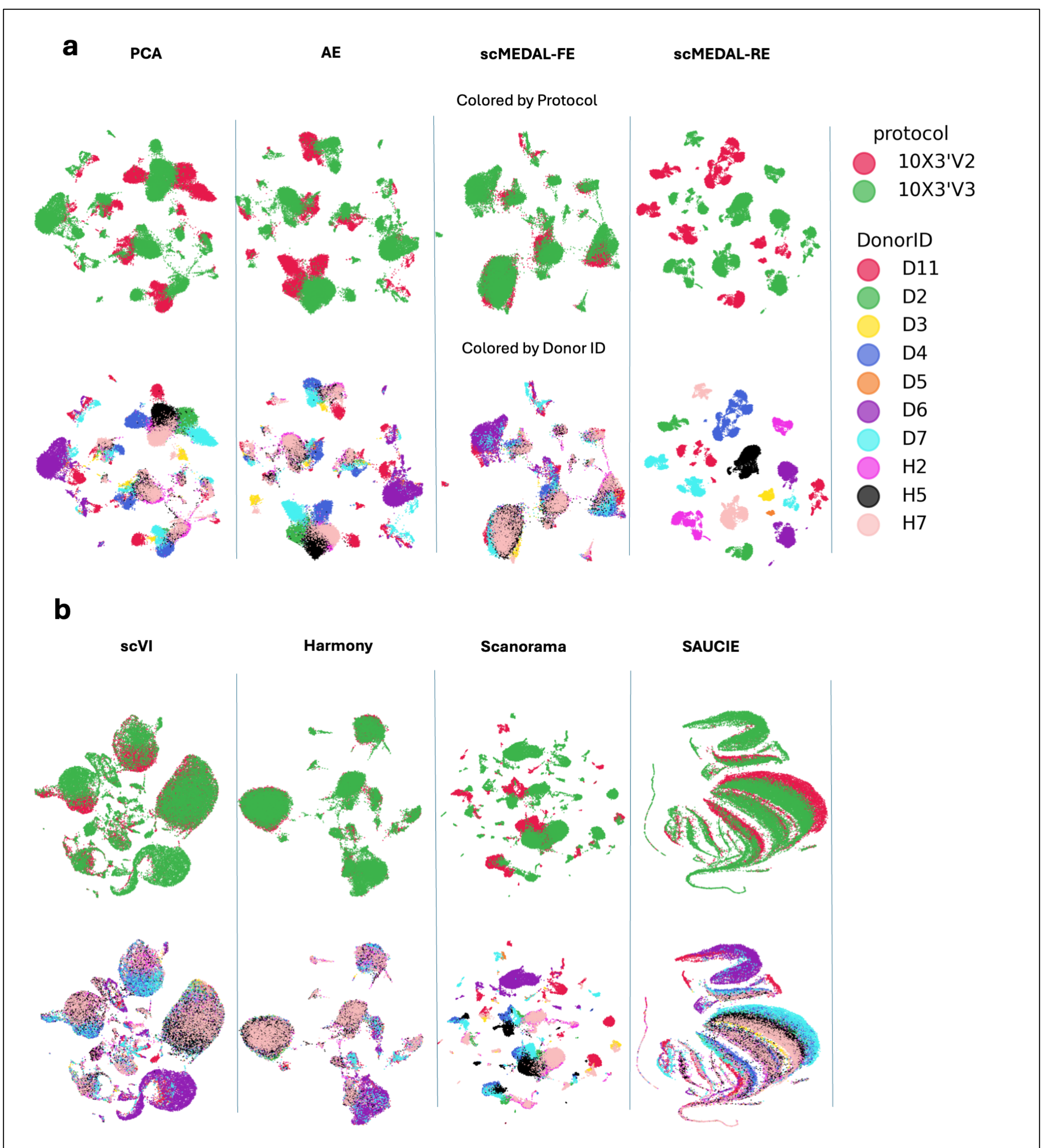

**Fig. S3: scMEDAL-RE separates cells from different protocols and donors in the Healthy Heart dataset. a** UMAP visualization of 41544 cells from 20 out of 147 batches of the Healthy Heart dataset's latent spaces colored by protocol (Chromium Single Cell 3′ v2 and v3; 10X3′V2 and 10X3′V3) and donor ID, generated using the following models: PCA (first column), AE (second column), scMEDAL-FE (third column) and scMEDAL-RE (fourth column) and in **b** scVI (first column), Harmony (second column), Scanorama (third column) and SAUCIE (fourth column).

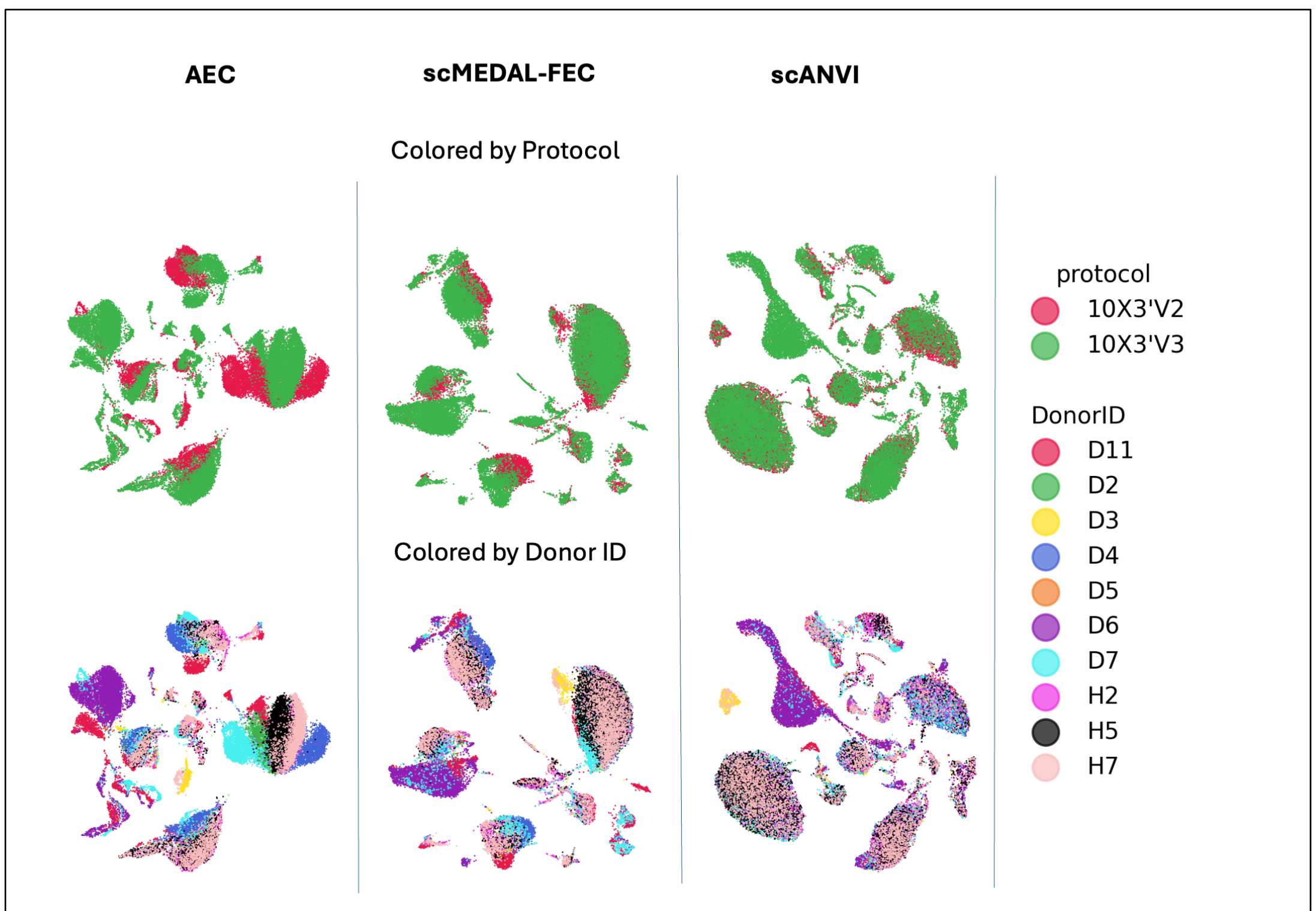

**Fig. S4**: **Supervised autoencoders show different degree of cell mixing between protocols.** UMAP visualization of AEC, scMEDAL-FEC and scANVI latent components colored by protocol (Chromium Single Cell 3′ v2 and v3; 10X3′V2 and 10X3′V3) and donor ID, representing 41544 cells from 20 selected batches (out of 147 total).

### 3.6.3. Training and validation curves

**Figs.** S5-S7 present the training and validation curves, for the AE, scMEDAL-FE, and scMEDAL-RE models in all three datasets. For scMEDAL-FE, the total loss is calculated as the reconstruction loss minus the adversarial loss, with each component scaled by its respective weight. In the scMEDAL-RE curves, the left axis displays the reconstruction loss and total loss, while the right axis shows: the latent cluster loss classifying a cell's batch and the Kullback-Leibler divergence (KLD) loss. Also shown is the total loss, which is computed as the sum of the reconstruction loss, latent cluster loss, and KLD loss, each adjusted by their respective weights. We observe that through all models, no one term dominates. Instead, the individual terms have values in similar ranges and can contribute effectively to the overall loss, thereby guiding the model fitting.

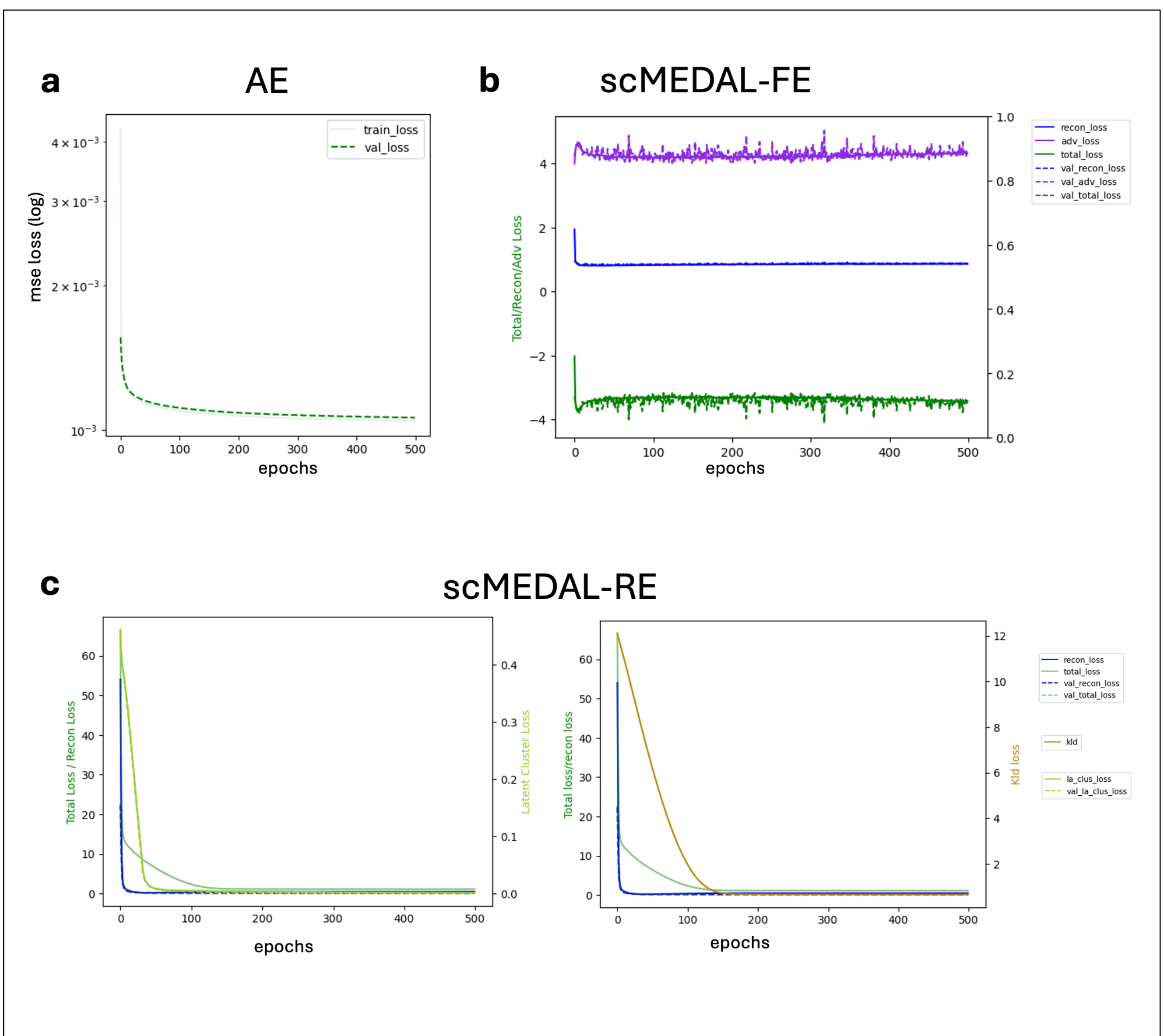

**Fig. S5: Training and validation curves for the Healthy Heart dataset.** Individual loss terms (see legend) maintain a similar scale throughout training. Training and validation curves adjusted by weights for **a** AE, **b** scMEDAL-FE (total loss = reconstruction loss - adversarial loss), and **c** scMEDAL-RE (total loss = reconstruction loss + KLD loss + latent cluster loss).

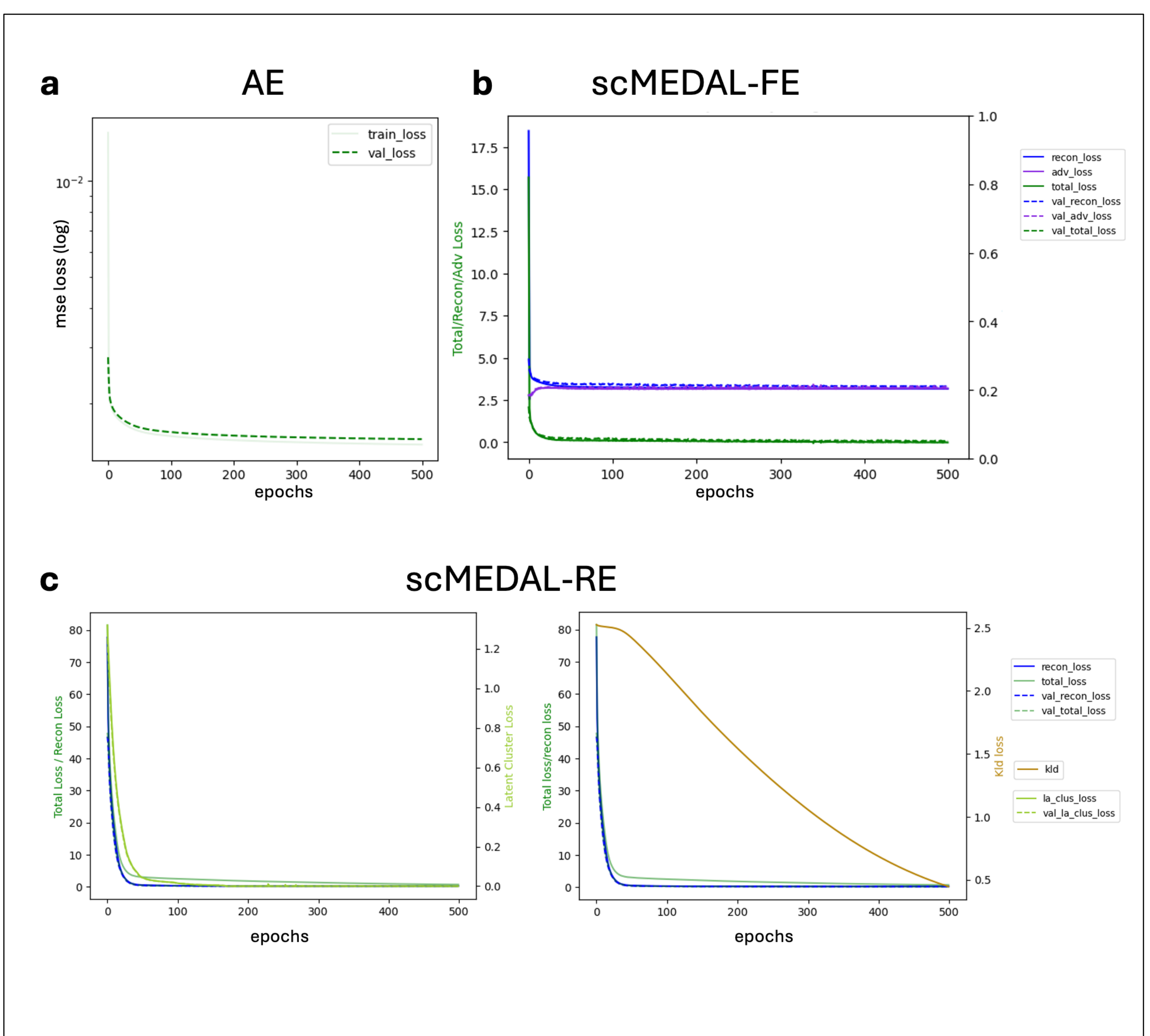

**Fig. S6: Training and validation curves for the ASD dataset.** Individual loss terms (see legend) maintain a similar scale throughout training. Training and validation curves adjusted by weights for **a** AE, **b** scMEDAL-FE (total loss = reconstruction loss - adversarial loss), and **c** scMEDAL-RE (total loss = reconstruction loss + KLD loss + latent cluster loss).

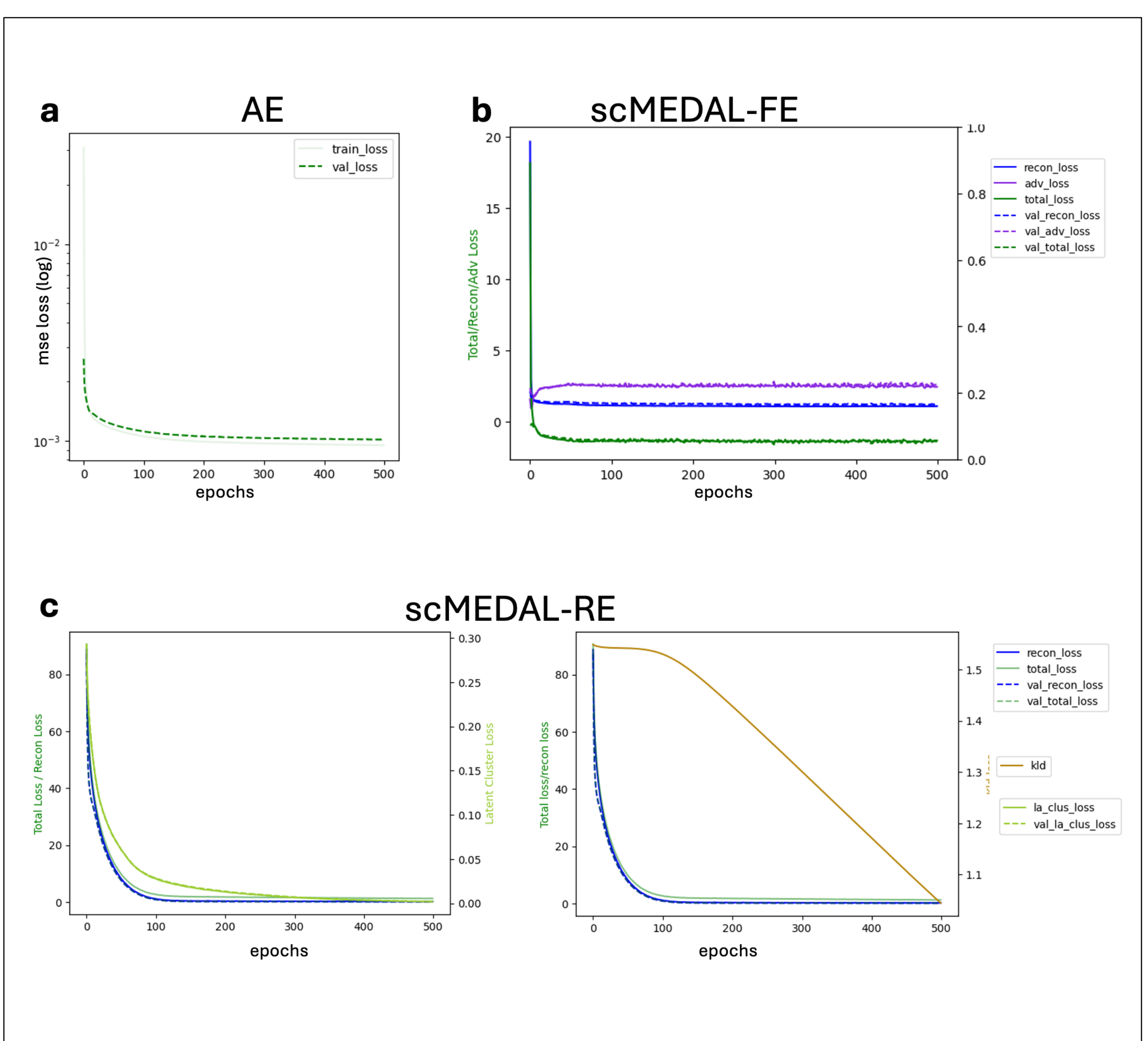

**Fig. S7: Training and validation curves for the AML dataset.** Individual loss terms (see legend) maintain a similar scale throughout training. Training and validation curves adjusted by weights for **a** AE, **b** scMEDAL-FE (total loss = reconstruction loss - adversarial loss), and **c** scMEDAL-RE (total loss = reconstruction loss + KLD loss + latent cluster loss).

## 3.7. Supplementary results for scMEDAL-FEC, AEC and scANVI models

### 3.7.1. CH and 1/DB scores

In **Tables** S9-S11 scores that complement the primary metric, ASW, are shown, including the Calinski-Harabasz (CH) clustering index and the reciprocal of the Davies-Bouldin (DB) clustering index (1/DB). These results further corroborate the ASW results in Section 2.6 of the main paper. As expected, AEC, scMEDAL-FEC and scANVI[8] show improvement in preserving cell type information compared with their respective non supervised comparable (**Tables** S6-S11). None of the models show a higher score for batch than scMEDAL-RE (**Tables** S9-S11). As described in the main paper (Section 2.6), the UMAP latent space visualizations show reduced separability in important biological information: tissue for Healthy Heart, diagnosis for ASD, patient group for AML.

**Table S9.** 1/DB and CH scores (Mean and 95% CI across 5 folds) for batch and cell type separability in the latent spaces of the **Healthy Heart dataset**, using AEC, scMEDAL-FEC and scANVI models. Bolded scores indicate the highest separability for each metric.

| | batch | | | | | | cell type | | | | | |
|---|---|---|---|---|---|---|---|---|---|---|---|---|
| | 1/DB | | | CH | | | 1/DB | | | CH | | |
| | mean | 95% CI | | mean | 95% CI | | mean | 95% CI | | mean | 95% CI | |
| AEC | 0.10 | 0.09 | 0.11 | 29.21 | 21.98 | 36.45 | 0.54 | 0.47 | 0.62 | 2473.06 | 1943.55 | 3002.56 |
| scMEDAL-FEC | 0.10 | 0.09 | 0.10 | 15.73 | 10.18 | 21.28 | 0.51 | 0.45 | 0.57 | 1672.46 | 1335.95 | 2008.97 |
| scANVI | 0.06 | 0.06 | 0.07 | 8.53 | 7.39 | 9.68 | 0.65 | 0.58 | 0.72 | 3510.81 | 2787.12 | 4234.49 |

**Table S10.** 1/DB and CH scores (Mean and 95% CI across 5 folds) for batch and cell type separability in the latent spaces of the **ASD dataset**, using AEC, scMEDAL-FEC and scANVI models. Bolded scores indicate the highest separability for each metric.

| | batch | | | | | | cell type | | | | | |
|---|---|---|---|---|---|---|---|---|---|---|---|---|
| | 1/DB | | | CH | | | 1/DB | | | CH | | |
| | mean | 95% CI | | mean | 95% CI | | mean | 95% CI | | mean | 95% CI | |
| AEC | 0.06 | 0.05 | 0.07 | 18.46 | 12.89 | 24.03 | 0.67 | 0.65 | 0.70 | 5739.65 | 4529.05 | 6950.24 |
| scMEDAL-FEC | 0.04 | 0.04 | 0.05 | 10.76 | 8.77 | 12.75 | 0.55 | 0.45 | 0.66 | 4397.63 | 3736.00 | 5059.25 |
| scANVI | 0.06 | 0.06 | 0.07 | 15.56 | 14.67 | 16.46 | 1.13 | 1.05 | 1.21 | 6512.85 | 5885.91 | 7139.78 |

**Table S11.** 1/DB and CH scores (Mean and 95% CI across 5 folds) for batch and cell type separability in the latent spaces of the **AML dataset**, using AEC, scMEDAL-FEC and scANVI models. Bolded scores indicate the highest separability for each metric.

| | batch | | | | | | cell type | | | | | |
|---|---|---|---|---|---|---|---|---|---|---|---|---|
| | 1/DB | | | CH | | | 1/DB | | | CH | | |
| | mean | 95% CI | | mean | 95% CI | | mean | 95% CI | | mean | 95% CI | |
| AEC | 0.13 | 0.11 | 0.15 | 118.27 | 105.19 | 131.35 | 0.33 | 0.31 | 0.36 | 1479.67 | 1195.23 | 1764.10 |
| scMEDAL-FEC | 0.11 | 0.10 | 0.12 | 116.67 | 86.30 | 147.03 | 0.27 | 0.23 | 0.32 | 1340.80 | 870.31 | 1811.29 |
| scANVI | 0.12 | 0.12 | 0.13 | 50.41 | 45.82 | 55.00 | 0.61 | 0.58 | 0.64 | 1449.83 | 1397.16 | 1502.50 |

### 3.7.2. Average training and validation curves for AEC and scMEDAL-FE across 5 folds in the Healthy Heart, ASD, and AML datasets

**Figs.** S8-S10 present the training and validation curves, for the AEC and scMEDAL-FEC models in the three datasets. For the AEC model, the total loss is calculated as the sum of the reconstruction loss and the cell type classification loss. For scMEDAL-FEC, the total loss is calculated as the sum of the reconstruction loss and the cell type classification loss minus the adversarial loss. We observe that in all cases the individual terms have similar value ranges and contribute effectively to the total loss.

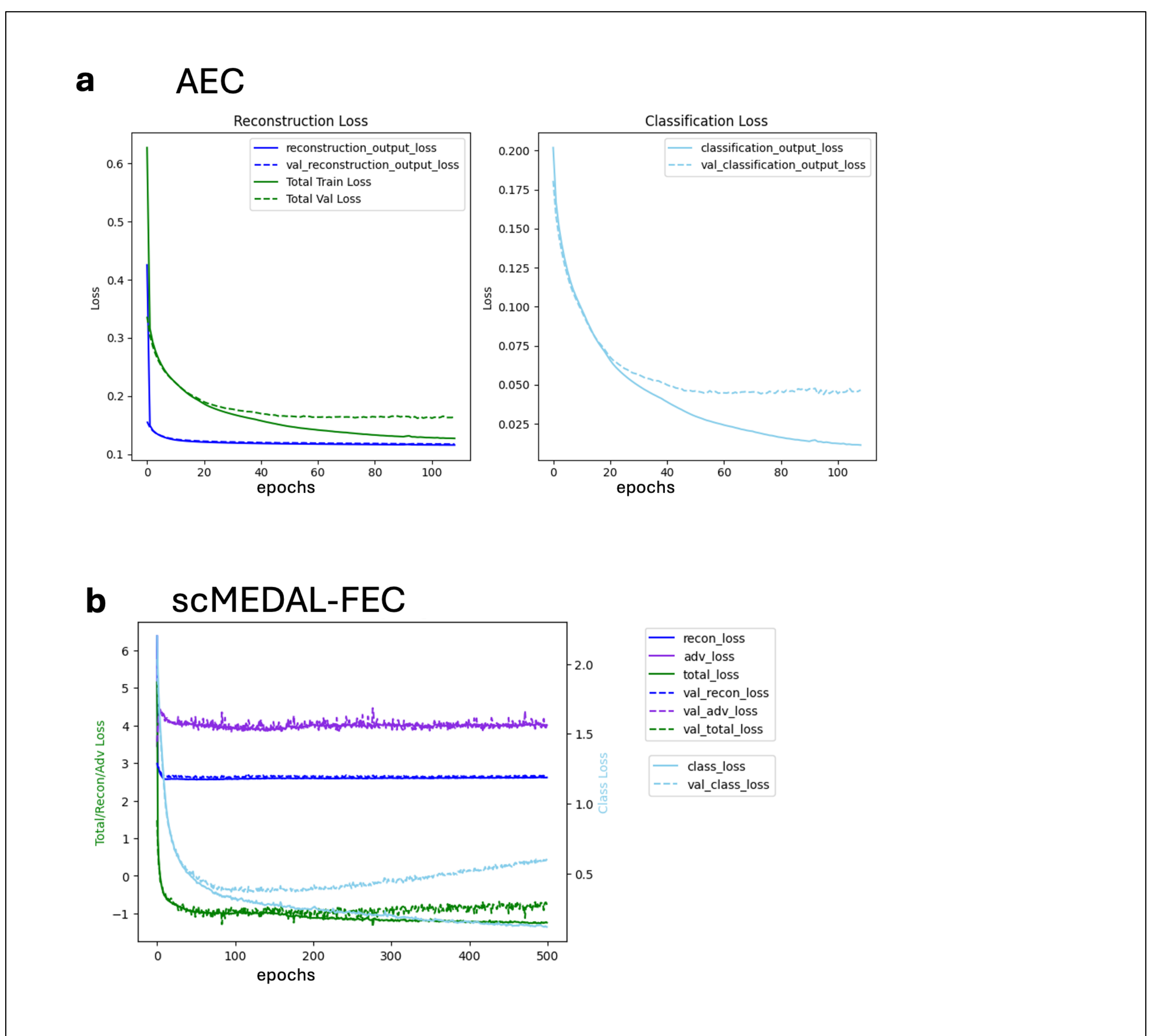

**Fig. S8: The training and validation curves for the Healthy Heart dataset demonstrate that the weight-adjusted individual loss terms effectively balance their contributions to the total loss.** Training and validation curves for the Healthy Heart dataset for **a** AEC and **b** scMEDAL-FEC models.

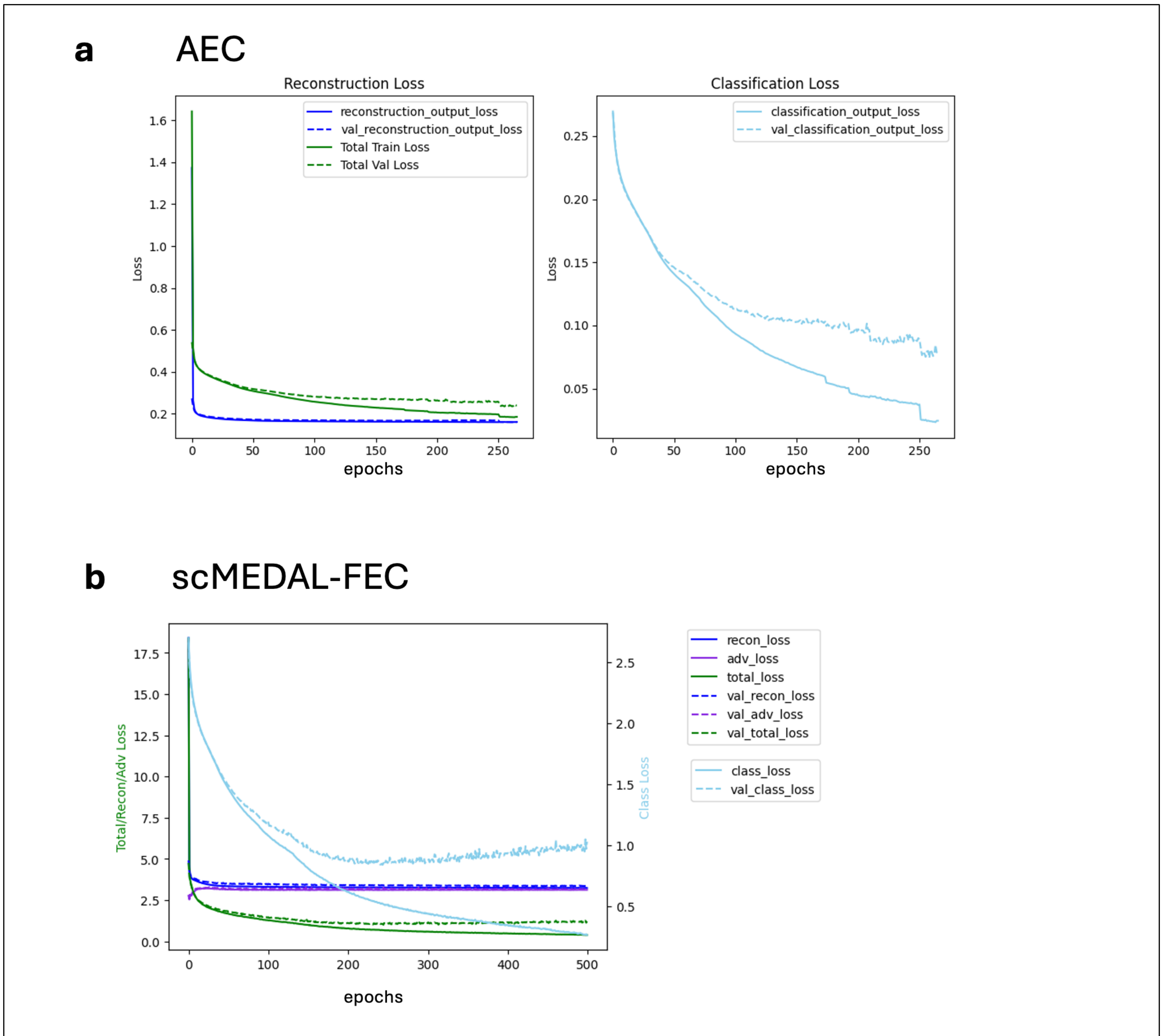

**Fig. S9: The training and validation curves for the ASD dataset demonstrate that the weight-adjusted individual loss terms effectively balance their contributions to the total loss.** Training and validation curves for the ASD dataset for **a** AEC and **b** scMEDAL-FEC models.

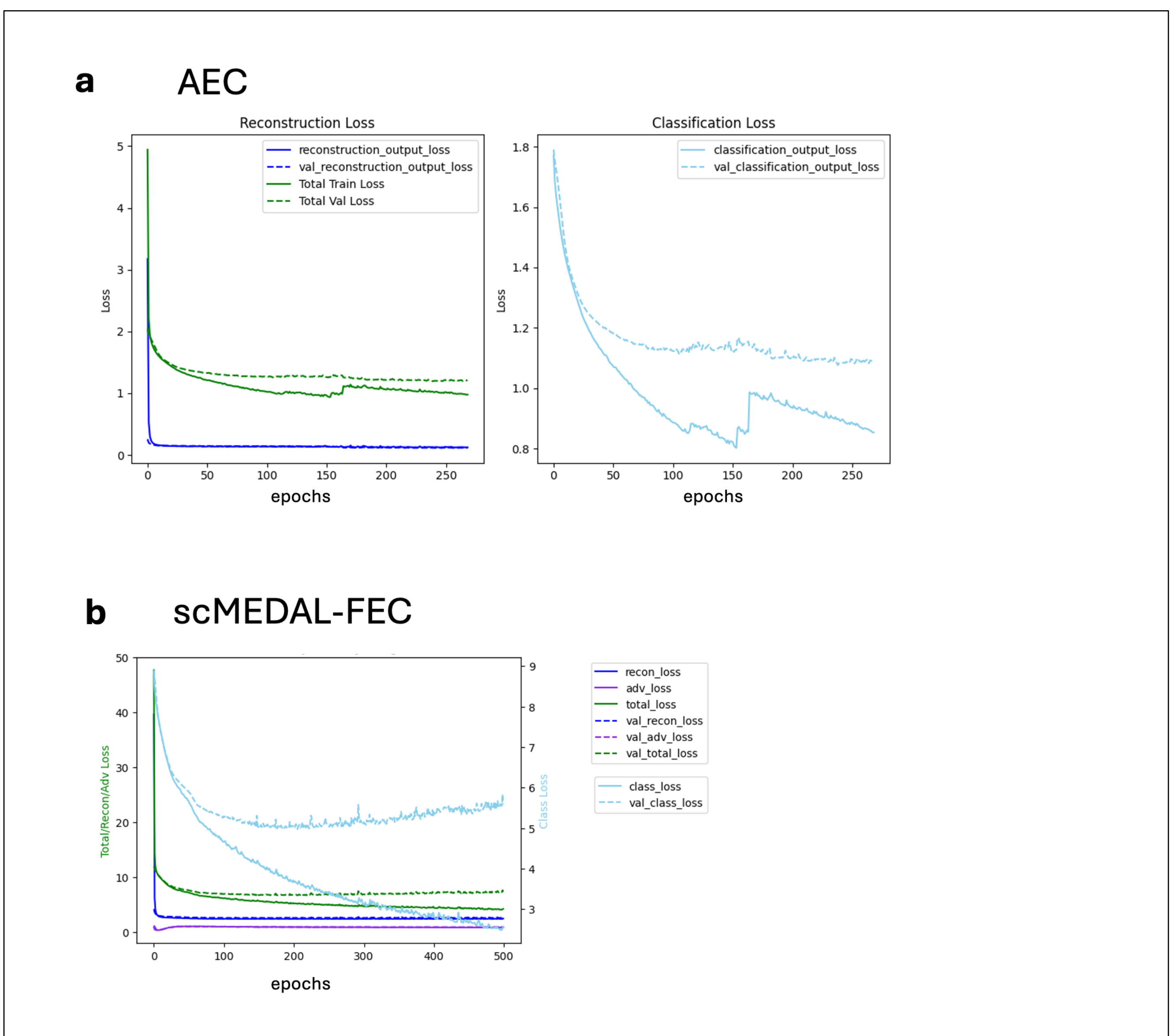

**Fig. S10: The training and validation curves for the AML dataset demonstrate that the weight-adjusted individual loss terms effectively balance their contributions to the total loss.** Training and validation curves for the AML dataset for **a** AEC and **b** scMEDAL-FEC models.

## 3.8. Mixed Effects classifier results using cell type as target

**Table** S12 shows the results for predicting cell-type using as input of a Random Forest classifier, combined scMEDAL-RE with latent spaces of Harmony, PCA, SAUCIE, Scanorama, scMEDAL-FE and scVI. AML shows the higher improvement when adding scMEDAL-RE, likely due to its strong cell-type and batch confounding.

**Table S12. Combining various embeddings with batch-specific (scMEDAL-RE) latent spaces improves the accuracy of cellular level predictions for targets that contain biological information.** Mean results across 5 folds and 95% confidence intervals (CI) for a Random Forest Classifier using tissue (Healthy Heart), diagnosis (ASD), and patient group (AML) as targets. The performance metrics include accuracy, balanced accuracy, and chance accuracy. Boldface denotes cases where pairing a model with scMEDAL-RE yields higher balanced accuracy, and the combined model also surpasses the standalone model in overall accuracy.

| | | | Accuracy | | | Balanced Accuracy | | | Chance Accuracy | | |
|---|---|---|---|---|---|---|---|---|---|---|---|
| **Dataset** | **Target** | **Latent space** | **mean** | **95%CI** | | **mean** | **95%CI** | | **mean** | **95%CI** | |
| **HH n=486,134** | Cell type (k=13) | Harmony | 89.1 | 89 | 89.19 | 73.72 | 73.25 | 74.18 | 16.2 | 16.11 | 16.29 |
| | | Harmony+scMEDAL-RE | 93.89 | 93.8 | 93.99 | **81.88** | **81.57** | **82.18** | | | |
| | | PCA | 94.52 | 94.45 | 94.6 | 83.97 | 83.62 | 84.32 | | | |
| | | PCA+scMEDAL-RE | 94.39 | 94.33 | 94.46 | 83.61 | 83.33 | 83.9 | | | |
| | | SAUCIE | 91.6 | 91.15 | 92.05 | 80.82 | 80.03 | 81.61 | | | |
| | | SAUCIE+scMEDAL-RE | 92.36 | 91.89 | 92.83 | **81.06** | **80.02** | **82.1** | | | |
| | | Scanorama | 93.93 | 93.88 | 93.98 | 82.47 | 82.13 | 82.8 | | | |
| | | Scanorama+scMEDAL-RE | 93.96 | 93.9 | 94.02 | **82.46** | **82** | **82.92** | | | |
| | | scMEDAL-FE | 91.6 | 91.37 | 91.82 | 77.68 | 76.46 | 78.91 | | | |
| | | scMEDAL-FE+scMEDAL-RE | 92.61 | 92.42 | 92.8 | **79.04** | **77.87** | **80.21** | | | |
| | | scVI | 92.44 | 92.06 | 92.83 | 83 | 82.49 | 83.51 | | | |
| | | scVI+scMEDAL-RE | 94.01 | 93.86 | 94.17 | **84.78** | **84.36** | **85.19** | | | |
| **ASD n=104,559** | Cell type (k=17) | Harmony | 92.15 | 92.01 | 92.29 | 90.41 | 90.22 | 90.59 | 7.86 | 7.76 | 7.96 |
| | | Harmony+scMEDAL-RE | 93.23 | 92.99 | 93.47 | **91.56** | **91.2** | **91.92** | | | |
| | | PCA | 93.62 | 93.41 | 93.82 | 92.19 | 91.91 | 92.47 | | | |
| | | PCA+scMEDAL-RE | 93.7 | 93.54 | 93.86 | 92.16 | 91.89 | 92.44 | | | |
| | | SAUCIE | 91.39 | 90.75 | 92.02 | 89.52 | 88.72 | 90.31 | | | |
| | | SAUCIE+scMEDAL-RE | 92.5 | 92.08 | 92.92 | **90.74** | **90.15** | **91.32** | | | |
| | | Scanorama | 91.54 | 91.12 | 91.97 | 89.47 | 88.99 | 89.96 | | | |
| | | Scanorama+scMEDAL-RE | 92.63 | 92.39 | 92.86 | **90.69** | **90.31** | **91.07** | | | |
| | | scMEDAL-FE | 92.09 | 91.78 | 92.4 | 89.66 | 89.17 | 90.15 | | | |
| | | scMEDAL-FE+scMEDAL-RE | 92.7 | 92.43 | 92.98 | **90.4** | **89.94** | **90.85** | | | |
| | | scVI | 90.31 | 90.02 | 90.6 | 88.21 | 87.55 | 88.86 | | | |
| | | scVI+scMEDAL-RE | 92.45 | 91.89 | 93 | **90.45** | **89.56** | **91.34** | | | |
| **AML n=38,417** | Cell type (k=21) | Harmony | 63.58 | 62.47 | 64.68 | 49.24 | 47.49 | 51 | 8 | 7.92 | 8.07 |
| | | Harmony+scMEDAL-RE | 79.56 | 79.26 | 79.86 | **71.42** | **70.75** | **72.08** | | | |
| | | PCA | 78.87 | 78.06 | 79.67 | 70.32 | 68.87 | 71.77 | | | |
| | | PCA+scMEDAL-RE | 79.73 | 79.28 | 80.17 | **71.96** | **70.96** | **72.97** | | | |
| | | SAUCIE | 74.98 | 74.21 | 75.75 | 68.43 | 67.35 | 69.51 | | | |
| | | SAUCIE+scMEDAL-RE | 78.82 | 78.14 | 79.5 | **72.69** | **71.69** | **73.68** | | | |
| | | Scanorama | 71.01 | 70.11 | 71.91 | 60.65 | 59.51 | 61.78 | | | |
| | | Scanorama+scMEDAL-RE | 78.7 | 78.05 | 79.36 | **70.74** | **69.86** | **71.63** | | | |
| | | scMEDAL-FE | 76.73 | 75.85 | 77.61 | 67.56 | 65.83 | 69.29 | | | |
| | | scMEDAL-FE+scMEDAL-RE | 78.87 | 78.49 | 79.25 | **70.77** | **70.21** | **71.33** | | | |
| | | scVI | 72.25 | 71.46 | 73.04 | 66.88 | 65.96 | 67.81 | | | |
| | | scVI+scMEDAL-RE | 79.15 | 79.05 | 79.25 | **73.82** | **73.42** | **74.23** | | | |

## 3.9. Significant genes from the genomaps for AML *versus* controls and ASD *versus* controls

Differential expression was assessed with a Statsmodels[18] linear mixed-effects model (LME) applied gene-wise to 300 Genomap-projected L2/3 cells, with diagnostic group (Typically Developing Control vs. ASD) as the fixed effect and a random intercept for donor batch (15 ASD, 16 control). Genes were called significant when the p-value for the group coefficient ($p_{\beta_1\, slope}$) was < 0.05. With treatment coding that sets ASD as the reference, the intercept $\beta_0$ estimates the ASD mean expression conditional on batch; its p-value tests $H_0\!: \beta_0 = 0$. The group coefficient $\beta_1$ represents the contrast $Control - ASD$ and its p-value tests $H_0\!: \beta_1 = 0$ . Positive $\beta_1$ indicates lower expression in ASD (higher in Control), whereas negative $\beta_1$ indicates higher expression in ASD. A total of 196 genes were identified as significantly associated with ASD ($p_{\beta_1 slope} < 0.05$). Statistics for these ASD-relevant genes are illustrated in **Table** S13. The full list of all ASD mapped genomap genes is available in Supplementary Data 1.

**Table S13.** Statistics for relevant genes associated with ASD. These genes are identified as significant through linear mixed-effects modeling comparing scMEDAL-RE projections of the same 300 L2/3 cells across 15 ASD and 16 control donors. Bolded scores highlight significant LME coefficient $\beta_1(\mathrm{Control} - \mathrm{ASD})$ together with the associated p-val.

| genes | pixel_i | pixel_j | $\beta_0$ intercept | 95% CI ($\beta_0$ intercept) | | p-val $\beta_0$ intercept | $\beta_1$ slope (control) | 95% CI ($\beta_1$ slope) | | p-val $\beta_1$ slope |
|---|---|---|---|---|---|---|---|---|---|---|
| ATP1A3 | 18 | 5 | 0.1176 | 0.0363 | 0.1989 | 0.0046 | **-0.2188** | **-0.3359** | **-0.1018** | **0.0002** |
| ATP1A2 | 32 | 50 | 0.1761 | -0.0388 | 0.3910 | 0.1083 | **-0.3204** | **-0.6297** | **-0.0111** | **0.0423** |
| FRYL | 36 | 48 | 0.1099 | -0.0134 | 0.2333 | 0.0806 | **-0.2461** | **-0.4236** | **-0.0686** | **0.0066** |
| EOMES | 37 | 28 | -0.5012 | -0.9296 | -0.0729 | 0.0218 | **0.7734** | **0.1568** | **1.3900** | **0.0140** |
| CYFIP1 | 50 | 37 | -0.1287 | -0.2918 | 0.0345 | 0.1222 | **0.2941** | **0.0592** | **0.5289** | **0.0141** |
| KCNJ2 | 53 | 8 | 0.1271 | -0.0614 | 0.3156 | 0.1864 | **-0.2978** | **-0.5691** | **-0.0264** | **0.0315** |

**Table** S14 presents the statistics for significant genes related to AML. These are identified by a Mann-Whitney U test performed on the same set of 300 monocytes projected with scMEDAL-RE onto 12 AML donors and then onto 5 healthy control subjects. Averages from the AML subjects and the control (healthy) subjects were computed for each set of 300 monocytes, and *stats* module from SciPy[17]was used to conduct the Mann-Whitney U test to identify gene expression differences between these groups. In total, 378 genes were found to have significantly different expression levels ($p < 0.05$). The full list of all AML mapped genomap genes is available in Supplementary Data 1.

**Table S14.** Statistics for relevant genes associated with AML, identified as significant through a Mann-Whitney U test on the same 300 monocyte Genomap projections from 12 AML scMEDAL-RE analyses and 5 control scMEDAL-RE analyses.

| genes | pixel_i | pixel_j | p-val |
|---|---|---|---|
| APC | 4 | 40 | 0.0027 |
| AFF3 | 3 | 34 | 0.0080 |
| FAT1 | 4 | 29 | 0.0080 |
| PRDM1 | 16 | 25 | 0.0127 |
| FTL | 32 | 2 | 0.0193 |
| FPR1 | 50 | 31 | 0.0280 |
| NFATC2 | 52 | 17 | 0.0280 |
| SETBP1 | 29 | 23 | 0.0280 |